%% file: main.tex
\documentclass[10pt,twocolumn,letterpaper]{article}

\usepackage[pagenumbers]{cvpr} 

\usepackage{graphicx}
\usepackage{amsmath}
\usepackage{amssymb}
\usepackage{booktabs}
\usepackage{enumitem}
\usepackage{soul}
\usepackage{array}
\usepackage{bbm}
\usepackage{multirow}
\usepackage[table,dvipsnames]{xcolor}
\usepackage{pifont}
\usepackage{tocloft}
\usepackage{titletoc}
\definecolor{hlrowcolor2}{RGB}{253,246,227}  
\definecolor{hlrowcolor}{RGB}{220,227,242}
\newcolumntype{L}[1]{>{\raggedright\let\newline\\\arraybackslash\hspace{0pt}}m{#1}}
\newcolumntype{C}[1]{>{\centering\let\newline\\\arraybackslash\hspace{0pt}}m{#1}}
\newcolumntype{R}[1]{>{\raggedleft\let\newline\\\arraybackslash\hspace{0pt}}m{#1}}
\newcommand*{\affmark}[1][*]{\textsuperscript{#1}}
\definecolor{urlcolor}{rgb}{0.93,0.01,0.55}

\usepackage[pagebackref,breaklinks,colorlinks]{hyperref}

\usepackage[capitalize]{cleveref}
\crefname{section}{Sec.}{Secs.}
\Crefname{section}{Section}{Sections}
\Crefname{table}{Table}{Tables}
\crefname{table}{Tab.}{Tabs.}

\input{math_commands.tex}

\usepackage{url}
\usepackage{algorithm}
\usepackage[]{algpseudocode}
\usepackage{pifont}

\newcommand{\cmark}{\ding{51}}
\newcommand{\xmark}{\ding{55}}
\definecolor{myred}{RGB}{220,50,47} 
\definecolor{mygreen}{RGB}{133,153,0}
\definecolor{commentcolor}{RGB}{133,153,0}
\newcommand{\greencmark}{\textcolor{mygreen}{\cmark}}
\newcommand{\redxmark}{\textcolor{myred}{\xmark}}
\definecolor{mygreen2}{RGB}{202,223,183}
\definecolor{myblue2}{RGB}{176,198,221}
\definecolor{myblue3}{RGB}{176,198,221}

\def\genbox#1#2#3#4#5#6{
    \leavevmode\raise#4bp\hbox to#5bp{\vrule height#5bp depth0bp width0bp
    \pdfliteral{q .5 w \csname #2COLOR\endcsname\space RG
                       \csname #3PDF\endcsname{#5}{#6} S Q
             \ifx1#1 q \csname #2COLOR\endcsname\space rg 
                       \csname #3PDF\endcsname{#5}{#6} f Q\fi}\hss}}

\newcommand{\comment}[1]{}
\cftsetindents{section}{0em}{1.7em}
\cftsetindents{subsection}{0em}{2.4em}

\def\cvprPaperID{***} 
\def\confName{CVPR}
\def\confYear{2022}

\begin{document}

\title{{Show Me What and Tell Me How: Video Synthesis via Multimodal Conditioning}}

\author{Ligong Han\affmark[1]\affmark[2]\thanks{Work done during an internship at Snap Inc. }\quad\quad
\quad Jian Ren\affmark[1]\quad\quad Hsin-Ying Lee\affmark[1] \quad\quad
Francesco Barbieri\affmark[1]\quad\quad \\ Kyle Olszewski\affmark[1]\quad\quad Shervin Minaee\affmark[1] \quad\quad Dimitris Metaxas\affmark[2] \quad\quad Sergey Tulyakov\affmark[1]\\
{\affmark[1]Snap Inc.\quad\quad\quad\affmark[2]Rutgers University }
}
\maketitle

\begin{abstract}
Most methods for conditional video synthesis use a single modality as the condition. 
This comes with major limitations. For example, it is problematic for a model conditioned on an image to generate a specific motion trajectory desired by the user since there is no means to provide motion information. Conversely, language information can describe the desired motion, while not precisely defining the content of the video. This work presents a multimodal video generation framework that benefits from text and images provided jointly or separately. We leverage the recent progress in quantized representations for videos and apply a bidirectional transformer with multiple modalities as inputs to predict a discrete video representation. To improve video quality and consistency, we propose a new video token trained with self-learning and an improved mask-prediction algorithm for sampling video tokens. We introduce text augmentation to improve the robustness of the textual representation and diversity of generated videos. Our framework can incorporate various visual modalities, such as segmentation masks, drawings, and partially occluded images. It can generate much longer sequences than the one used for training. In addition, our model can extract visual information as suggested by the text prompt, \emph{e.g.}, ``an object in image one is moving northeast'', and generate corresponding videos. We run evaluations on three public datasets and a newly collected dataset labeled with facial attributes, achieving state-of-the-art generation results on all four\footnote{Code:~\href{https://github.com/snap-research/MMVID}{\color{urlcolor}{https://github.com/snap-research/MMVID}} and~\href{https://snap-research.github.io/MMVID/}{\color{urlcolor}{Webpage}}.}.
\end{abstract}

\section{Introduction}
Generic video synthesis methods generate videos by sampling from a random distribution~\cite{tulyakov2017mocogan,tian2021a}. To get more control over the generated content, conditional video synthesis works utilize input signals, such as images~\cite{Blattmann_2021_CVPR,Dorkenwald_2021_CVPR}, text or language~\cite{balaji2019conditional,kim2020tivgan}, and action classes~\cite{vondrick2016generating}. This enables synthesized videos containing the desired objects as specified by visual information or desired actions as specified by textual information.

Existing works on conditional video generation use only one of the possible control signals as inputs~\cite{tfgan,kim2020tivgan}. This limits the flexibility and quality of the generative process.
For example, given a screenplay we could potentially generate several movies, depending on the decisions of the director, set designer, and visual effect artist. 
In a similar way, a video generation model conditioned with a text prompt should be primed with different visual inputs.
Additionally, a generative video model conditioned on a given image should be able to learn to generate various plausible videos, which can be defined from various natural language instructions.
For example, to generate object-centric videos with objects moving~\cite{wu2021generative}, the motion can be easily defined through a text prompt, \textit{e.g.}, ``moving in a zig-zag way,'' while the objects can be defined by visual inputs. 
Thus, an interesting yet challenging question arises: 
\textit{Can we learn a video generation model that can support such behavior?} 

We tackle the question in this work and propose a new video synthesis model supporting diverse, multimodal conditioning signals. Our method consists of two phases. The \emph{first} phase obtains discrete representations from images.
We employ an autoencoder with a quantized bottleneck, inspired by the recent success of two-stage image generation using quantized feature representations~\cite{oord2017neural,razavi2019generating,esser2020taming,zhang2021ufc}. 
The \emph{second} phase learns to generate video representations that are conditioned on the input modalities, which can then be decoded into videos using the decoder from the first stage. 
We leverage a bidirectional transformer, \emph{i.e.}, BERT~\cite{devlin2018bert}, trained with a masked sequence modeling task, that uses tokens from multimodal samples and predicts the latent representation for videos. Building such a framework requires solving several challenging problems. First, video consistency is a common problem among video generation methods. 
Second, it is necessary to ensure that the correct textual information is learned.
Third, training a transformer for image synthesis is computationally demanding~\cite{chen2020generative}, an issue that is even more severe in the time domain, as a longer sequence of tokens needs to be learned. To solve these challenges, we propose the following contributions: 
\begin{itemize}[leftmargin=1.5em]
  \setlength\itemsep{-0.25em}
  \item We introduce a bidirectional transformer with several new techniques to improve video generation: For training, we propose the video token \texttt{VID}, which is trained via self-learning and video attention, to model temporal consistency; For inference, we improve mask-predict to generate videos with improved quality.
  \item 
  We introduce text augmentation, including text dropout and pretrained language models for extracting textual embeddings, to generate diverse videos that are correlated with the provided text.
  \item  We explore long sequence synthesis with the transformer model to generate sequences with lengths that are much longer than the one used for training (Fig.~\ref{fig:iper}).
\end{itemize}

We name our framework \textbf{MMVID} and show that a \textbf{M}ulti\textbf{M}odal \textbf{VID}eo generator can enable various applications. The user can show \emph{what} to generate using visual modalities and tell \emph{how} to generate with language.
We explore two settings for multimodal video generation. The first involves \textit{independent} multimodalities, such that there is no relationship between textual and visual controls (Fig.~\ref{fig:shape_vc} and Fig.~\ref{fig:vox_vc}). The second one targets \textit{dependent} multimodal generation, where we use text to obtain certain attributes from given visual controls (Fig.~\ref{fig:shape_text_dep} and Fig.~\ref{fig:vox_vc}). The latter case allows for more potential applications, in which language is not able to accurately describe certain image content that the user seeks to generate, but images can efficiently define such content. We also show our model can use diverse visual information, including segmentation masks, drawings, and partially observed images (Fig.~\ref{fig:vox_vc}).

To validate our approach extensively, we conduct experiments on \emph{four} datasets. In addition to three public datasets, we collect a new dataset, named Multimodal VoxCeleb, that includes $19,522$ videos from VoxCeleb~\cite{nagrani2020voxceleb} with $36$ manually labeled facial attributes.

\section{Related Works}

\noindent\textbf{Video Generation}. For simplicity, previous works on video generation can be categorized into unconditional and conditional generation, where most of them apply similar training strategies: adversarial training with image and video discriminators~\cite{goodfellow2014generative,clark2019adversarial,TGAN2020}.
Research on \textit{unconditional} video generation studies how to synthesize diverse videos with input latent content or motion noise~\cite{vondrick2016generating,saito2017temporal,tulyakov2017mocogan,yushchenko2019markov,acharya2018towards,wang2020g3an,hyun2021self}. Recent efforts in this direction have achieved high resolution and high quality generation results for images and videos~\cite{kahembwe2020lower,tian2021a,clark2019adversarial}. 
On the other hand, \textit{conditional} video generation utilizes given visual or textual information for video synthesis~\cite{pan2019video,wang2020imaginator,walker2021predicting,Blattmann_2021_CVPR,Dorkenwald_2021_CVPR,ren2021flow,siarohin2021motion}. For example,
the task of video prediction uses the provided first image or a few images to generate a sequence of frames~\cite{babaeizadeh2021fitvid,denton2017unsupervised,denton2018stochastic,walker2017pose,villegas17hierchvid,villegas17mcnet,babaeizadeh2017stochastic,hsieh2018learning,byeon2018contextvp}. Similarly, text-to-video generation applies the conditional signal from text,  captions, or natural language descriptions~\cite{li2018video,marwah2017attentive,pan2017create}. TFGAN~\cite{balaji2019conditional} proposes a  multi-scale text-filter conditioning scheme for discriminators. 
TiVGAN~\cite{kim2020tivgan} proposes to  generate a single image from text and synthesizes consecutive frames through further stages.
In this work, we study conditional video synthesis. However, we differ from existing methods since we address a more challenging problem: multimodal video generation. Instead of using a single modality, such as textual guidance, we show how multiple modalities can be input within a single  framework for video generation. With multimodal controls, \emph{i.e.}, textual and visual inputs, we further enhance two settings for video generation: independent and dependent multimodal inputs, in which various applications can be developed.

\noindent\textbf{Transformers for Video Generation}. Transformer-based networks have shown promising and often superior performance not only in natural languages processing tasks~\cite{brown2020language,radford2018improving,vaswani2017attention}, but also in computer vision related efforts~\cite{chen2020generative,dosovitskiy2020image,radford2021learning,huang2020pixel,qi2020imagebert}. Recent works prvoide promising results on conditional image generation~\cite{esser2020taming,cho2020x}, text-to-image generation~\cite{ramesh2021zero,ding2021cogview}, video generation~\cite{wu2021generative,rakhimov2020latent,Weissenborn2020Scaling,yan2021videogpt}, and text-to-video synthesis~\cite{wu2021godiva} using transformers.
Unlike existing transformer-based video generation works that focus on autoregressive training, we apply a non-autoregressive generation pipeline with a bidirectional transformer~\cite{gu2017non,gao2019masked,ghazvininejad2019mask,wang2019bert,mansimov2019generalized}. Our work is inspired by M6-UFC~\cite{zhang2021ufc}, which utilizes the non-autoregressive training for multimodal image generation and produces more diverse image generation with higher quality. Building upon M6-UFC, we further introduce training techniques for multimodal video synthesis.

\begin{figure*}[t]
    \centering
    \includegraphics[width=0.95\linewidth]{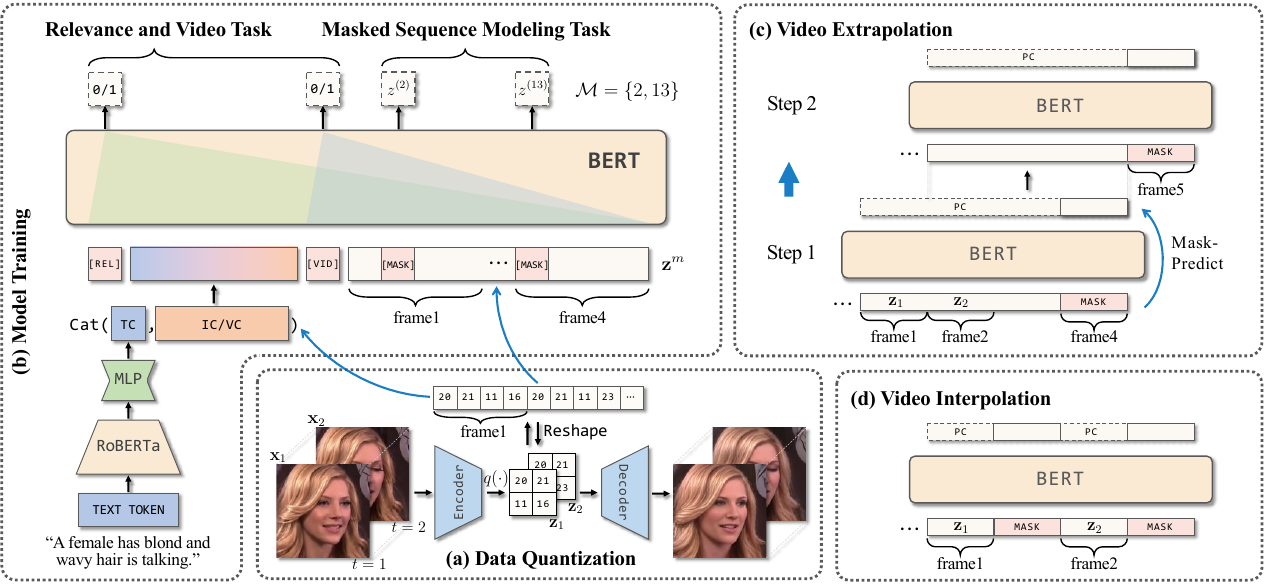}
    \caption{\textbf{Pipeline} for training and inference. (a) Data quantization. (b) Model training. Within the BERT module, the {green} and {blue} triangles indicate the attention scopes of \colorbox{mygreen2}{\texttt{[REL]}} and \colorbox{myblue2}{\texttt{[VID]}}, respectively. (c) Video extrapolation. For simplicity, each step represents a full mask-predict process instead of a single forward pass of the transformer. (d) Video interpolation.    }
    \label{fig:pipeline}
\end{figure*}

\section{Methods}
Our framework for multimodal video generation is a two-stage image generation method. It uses discrete feature representations~\cite{oord2017neural,razavi2019generating,esser2020taming,zhang2021ufc}. During the \emph{first} stage we train an autoencoder (with encoder $E$ and decoder $D$) that has the same architecture as the one from VQGAN~\cite{esser2020taming} to obtain a quantized representation for images. 
Given a real video clip defined as $\mathbf{v}=\{\mathbf{x}_1, \mathbf{x}_2, \cdots, \mathbf{x}_T\}$ with $\mathbf{x}_t\in\mathbb{R}^{H\times W\times 3}$, we get a quantized representation of the video defined as $\mathbf{z}=\{\mathbf{z}_1, \mathbf{z}_2, \cdots, \mathbf{z}_T\}$, where $\mathbf{z}_t = {q}(E(\mathbf{x}_t))\in\mathbb{N}_1^{h\times w}$. ${q}(\cdot)$ denotes the quantization operation and $\mathbb{N}_1$ indicates a set of positive integers.

During the \emph{second} stage we learn a bidirectional transformer for modeling the correlation between multimodal controls and the learned vector quantization representation of a video. 
Specifically, we concatenate the tokens from the multimodal inputs and the target video as a sequence to train the transformer.
Tensors obtained from an image and video must be vectorized for concatenation. We do so using the reshape operation ($\texttt{Reshape}$).
Therefore, we have a video tensor $\mathbf{z}$ reshaped into a single-index tensor as $\texttt{Reshape}(\mathbf{z})=[z^{(1)},\cdots,z^{(hwT)}]$.
For simplicity of notation, we define $\mathbf{z}\equiv\texttt{Reshape}(\mathbf{z})$.
To train the non-autogressive transformer (BERT) on video tokens, we employ three tasks: Masked Sequence Modeling (MSM),  RELevance estimation (REL), and VIDeo consistency estimation (VID). During inference, samples are generated via an iterative algorithm based on mask-predict~\cite{ghazvininejad2019mask}, which is simulated by the MSM task during training. The REL and VID tasks regularize the model to synthesize videos that are relevant to the multimodal signals and are temporally consistent.
In the following two sections we present each task.

\subsection{Masked Sequence Modeling with Relevance}\label{sec:msm} 
\noindent\textbf{Masked Sequence Modeling}. The MSM task is similar to a conditional masked language model~\cite{ghazvininejad2019mask}. It is essential for the non-autoregressive model to learn bidirectional representations and enables parallel generation (mask-predict). Inspired by M6-UFC~\cite{zhang2021ufc} and VIMPAC~\cite{tan2021vimpac}, we consider five masking strategies: (I) i.i.d. masking, \emph{i.e.}, randomly masking video tokens according to a Bernoulli distribution; 
(II) masking all tokens; (III) block masking~\cite{tan2021vimpac}, which masks continuous tokens inside spatio-temporal blocks; (IV) the negation of block masking, which preserves the spatio-temporal block and masks the rest of the tokens; (V) randomly keeping some frames (optional). Strategies I and II are designed to simulate mask-predict sampling (the strategy chosen most of the time). 
Strategy II helps the model learn to generate from a fully masked sequence in the first step of mask-predict. Strategies III - V can be used as Preservation  Control  (PC)  for preservation tasks, which enable the use of partial images as input (Figs.~\ref{fig:shape_vc},~\ref{fig:vox_vc}) and performing long sequence generation (Fig.~\ref{fig:iper}).
The MSM task minimizes the softmax cross-entropy loss $\mathcal{L}_\text{MSM}$ as follows:
\begin{equation}
\small
\mathcal{L}_\text{MSM}=-\frac{1}{|\mathcal{M}|}\sum_{i\in\mathcal{M}}{\log{P(z^{(i)}|{\mathbf{z}^m}, \mathbf{c})}},
\label{eq:msm}
\end{equation}
where $\mathcal{M}$ is the masking indices, ${\mathbf{z}^m}$ is the masked sequence, and $\mathbf{c}$ denotes the control sequence.

\noindent\textbf{Relevance Estimation}. To encourage the transformer to learn the correlation between multimodal inputs and target videos, we prepend a special token \texttt{REL} that is similar to the one used in M6-UFC~\cite{zhang2021ufc}  to the whole sequence, and learn a binary classifier to classify positive and negative sequences. The positive sequence is the same as the sequence used in the MSM task so that we can reuse the same transformer in the forward pass. The negative sequence is constructed by swapping the condition signals along the batch dimension. This swapping does not guarantee constructing strictly negative samples. Nevertheless, we still find it is good enough to make the model learn relevance in practice.
The loss function $\mathcal{L}_\text{REL}$ for the relevance task is given by:
\begin{equation}
\small
\mathcal{L}_\text{REL}=-\log{P(1|{\mathbf{z}^m}, \mathbf{c})}-\log{P(0|{\mathbf{z}^m}, \bar{\mathbf{c}})},
\label{eq:rel}
\end{equation}
where $\bar{\mathbf{c}}$ denotes the swapped control sequence.

\subsection{Video Consistency Estimation}
To further regularize the model to generate temporally consistent videos, we introduce the video consistency estimation task. Similar to \texttt{REL}, we use a special token \texttt{VID} to classify positive and negative sequences.

\noindent\textbf{Video Attention}. The VID task focuses on video token sequences. Thus, we place the \texttt{VID} token between the control and target sequences. We apply a mask to BERT to blind the scope of the \texttt{VID} token from the control signals so it only calculates attention from the tokens of the target videos. The positive sequence is the same one used in MSM and REL tasks. The negative sequence is obtained by performing negative augmentation on videos to construct samples that do not have temporally consistent motion or content.

\noindent\textbf{Negative Video Augmentation}. We employ four strategies to augment negative video sequences: (I) \emph{frame swapping} -- a random frame is replaced by using a frame from another video;
(II) \emph{frame shuffling} -- frames within a sequence are shuffled; 
(III) \emph{color jittering} -- randomly changing the color of one frame; (IV) \emph{affine transform} -- randomly applying an affine transformation on one frame. All augmentations are performed in image space.
With $\bar{\mathbf{z}}$ denoting the video sequence after augmentation, the loss $\mathcal{L}_\text{VID}$ for the VID task is:
\begin{equation}
\small
\mathcal{L}_\text{VID}=-\log{P(1|{\mathbf{z}^m}, \mathbf{c})}-\log{P(0|{\bar{\mathbf{z}}^m}, {\mathbf{c}})}.
\label{eq:vid}
\end{equation}
Overall, the full objective is $\mathcal{L}=\lambda_\text{MSM}\mathcal{L}_\text{MSM} + \lambda_\text{REL}\mathcal{L}_\text{REL} + \lambda_\text{VID}\mathcal{L}_\text{VID}$, where $\lambda$s balances the losses.

\subsection{Improved Mask-Predict for Video Generation}
We employ mask-predict~\cite{ghazvininejad2019mask} during inference, which iteratively remasks and repredicts low-confidence tokens by starting from a fully-masked sequence. We chose it because it 
can be used with our bidirectional transformer, as the length of the target sequence is fixed. In addition, mask predict provides several benefits. First, it allows efficient parallel sampling of tokens in a sequence. Second, the unrolling iterations from mask-predict enable direct optimization on synthesized samples, which can reduce exposure bias~\cite{ranzato2015sequence}. Third, information comes from both directions, which makes the generated videos more consistent.

We build our sampling algorithm based on the original mask-predict~\cite{ghazvininejad2019mask} with two improvements:
(I) noise-annealing multinomial sampling, \ie, adding noise during remasking; (II) a new scheme for mask annealing, \ie, using a piecewise linear annealing scheme to prevent the generated motion from being washed out after too many steps of mask-predict. We also apply a beam search from M6-UFC~\cite{zhang2021ufc}. In Alg.~\ref{alg:mask_predict}, the transformer ($\texttt{BERT}$) takes input tokens $\mathbf{z}_{in}$ and outputs score $s$ and the logits $\tilde{\mathbf{p}}$ for all target tokens. At each mask-predict iteration, we sample tokens with $\texttt{SampleToken}$ that returns a predicted token $\mathbf{z}_{out}$ and 
a vector $\mathbf{y}$ containing its probabilities (unnormalized).
$\texttt{SampleToken}$ also accepts a scalar $\sigma$ that indicates the noise level to be added during the token sampling process. $\texttt{SampleMask}(\mathbf{y},\mathbf{m},N-n)$ remasks $n$ tokens from a total of $N$ tokens according to the multinomial defined by the normalized $\mathbf{y}$, while ensuring tokens with $\mathbf{m}=1$ are always preserved. $\mathbf{z}_\phi$ denotes the fully-masked sequence. The functions $\texttt{SampleToken}$, $\texttt{SampleMask}$ and the schedules of $n^{(i)}$ and $\sigma^{(i)}$ are shown in Appendix.

\begin{algorithm}[h]
\caption{Improved Mask-Predict for Video Generation}
\label{alg:mask_predict}
\begin{algorithmic}[1]
    \Require Initial PC mask $\mathbf{m}_\text{PC}$ and initial token $\mathbf{z}_{in}$.
    \State $\tilde{\mathbf{p}}, s \leftarrow \texttt{BERT}(\mathbf{z}_{in})$
    \State $\mathbf{z}_{out}, \mathbf{y} \leftarrow \texttt{SampleToken}(\tilde{\mathbf{p}}, \sigma^{(1)})$
    \State $\mathbf{z}_{out} \leftarrow \mathbf{m}_\text{PC}\odot \mathbf{z}_{in} + (1-\mathbf{m}_\text{PC}) \odot  \mathbf{z}_{out}$ \Comment{\textcolor{commentcolor}{PC}}
    \For{$i \in \{2, ..., L\}$} \Comment{\textcolor{commentcolor}{main loop}}
        \For{$b \in \{1, ..., B\}$} \Comment{\textcolor{commentcolor}{beam search}}
            \State $\mathbf{m}^b \leftarrow \texttt{SampleMask}(\mathbf{y}, \mathbf{m}_\text{PC}, N-n^{(i)})$ 
            \State $\mathbf{z}^b_{in} \leftarrow \mathbf{m}^b\odot \mathbf{z}_{out} + (1-\mathbf{m}^b) \odot \mathbf{z}_{\phi}$ \Comment{\textcolor{commentcolor}{remask}}
            \State $\tilde{\mathbf{p}}^b, s^b \leftarrow \texttt{BERT}(\mathbf{z}^b_{in})$ \Comment{\textcolor{commentcolor}{repredict}}
        \EndFor
        \State $b^* \leftarrow \argmax_b(s^b)$
        \State $\mathbf{z}_{out}, \mathbf{y} \leftarrow \texttt{SampleToken}(\tilde{\mathbf{p}}^{b^*}, \sigma^{(i)})$
    \EndFor
    \State \Return $\mathbf{z}_{out}$
\end{algorithmic}
\end{algorithm}

\subsection{Text Augmentation}
We study two augmentation methods. First, we randomly drop sentences from the input text to avoid memorizing certain word combinations.
Second, we apply a fixed \emph{pretrained} language model, \emph{i.e.}, RoBERTa~\cite{liu2019roberta}, rather than learning text token embeddings in a lookup table from scratch, to let the model be more robust for input textual information.
The features of text tokens are obtained from an additional multilayer perceptron (MLP) appended after the language model that matches the vector dimension with BERT. 
The features are converted to a weighted sum to get the final embedding of the input text.
With the language model, the video generation framework is more robust for out-of-distribution text prompts. When using the tokenizer from an existing work~\cite{radford2021learning}, we observed that it might not properly handle synonyms without a common root (Fig.~\ref{fig:vox_lang}). 
\subsection{Long Sequence Generation}
Due to the inherent preservation control mechanism during training (strategy V in the MSM task), we can generate sequences with many more frames than the model is trained with via interpolation or extrapolation. \textbf{Interpolation} is conducted by generating the intermediate frames between given frames. As illustrated by Fig.~\ref{fig:pipeline} (d), we place $\mathbf{z}_1$ and $\mathbf{z}_2$ at the positions of frames 1 and 3 to serve as preservation controls, \ie, they are kept the same during mask-predict iterations, and we can interpolate a frame between them.
\textbf{Extrapolation} is similar to interpolation, except we condition the model on previous frames to generate the next frames. As illustrated in Fig.~\ref{fig:pipeline} (c), this process can be iterated a number of times to generate longer videos.

\section{Experiments}
\noindent \textbf{Datasets.} We show experiments on the following datasets.
\begin{itemize}[leftmargin=1em]
  \setlength\itemsep{-0.25em}
    \item \textbf{Shapes} is proposed by TFGAN~\cite{tfgan} for text-to-video generation. Each video shows one object (a geometric shape with specified color and size) displayed in a textured moving background. The motion of an object is described by a text and the background is moving in a random way. There are $30$K videos with size $64\times64$.
    \item \textbf{MUG}~\cite{aifanti2010mug} contains $52$ actors performing $6$ different facial expressions. 
    We also provide gender labels for the actors. For a fair comparison, we obtain text descriptions by following TiVGAN~\cite{kim2020tivgan}. We run experiments on $1039$ videos with resolution $128\times128$.
    \item \textbf{iPER}~\cite{liu2019liquid} consists of $206$ videos of $30$ subjects wearing different clothes performing an A-pose and random actions. Experiments are conducted with size $128\times128$.
    \item \textbf{Multimodal VoxCeleb} is a new dataset for multimodal video generation. We first obtain $19,522$ videos from  VoxCeleb~\cite{nagrani2020voxceleb} after performing  pre-processing~\cite{Siarohin_2019_NeurIPS}. Second, we manually label $36$ facial attributes described in CelebA~\cite{liu2015faceattributes} for each video. Third, we use a probabilistic context-free grammar to generate language descriptions~\cite{xia2021tedigan}. Finally, we run APDrawingGAN~\cite{YiLLR20} to get artistic portrait drawings and utilize face-parsing~\cite{yu2021bisenet} to produce segmentation masks. 
\end{itemize}

\noindent \textbf{Baseline Methods}. We run TFGAN~\cite{tfgan} on Shapes, MUG, and Multimodal VoxCeleb datases for comparison of text-to-video synthesis. We also compare our approach with TiVGAN~\cite{kim2020tivgan} on MUG. 
Additionally, we unify the autoregressive transformer of DALL-E~\cite{ramesh2021zero} and the autoencoder from VQGAN (the same one used in our method) in a multimodal video generative model. We name the strong baseline as {\bf A}uto{\bf R}egressive {\bf T}ransformer for {\bf V}ideo generation (\textbf{ART-V}) and compare it with our bidirectional transformer for predicting video tokens. 
We train ART-V with the next-token-prediction objective on concatenated token sequences obtained from input controls and target videos.

\noindent \textbf{Evaluation Metrics}. 
We follow the metrics from existing works on Shapes and MUG to get a fair comparison. Specifically, we compute classification accuracy on Shapes and MUG and Inception Score (IS)~\cite{salimans2016improved} on MUG. On Multimodal VoxCeleb and iPER datasets, we report Fr\'{e}chet Video Distances (FVD)~\cite{unterthiner2018towards} that is computed from $2048$ samples, and Precision-Recall Distribution (PRD) ($F_{8}$ and $F_{1/8}$) for diversity~\cite{precision_recall_distributions}. We further report CLIP score~\cite{radford2021learning} for calculating the cosine similarity between textual inputs and the generated videos on Multimodal VoxCeleb. 
\begin{figure}[t]
\centering
\begin{subfigure}[t] {1\linewidth}
    \centering
    \setlength{\tabcolsep}{0\linewidth}
    \begin{tabular}{L{1\linewidth}}
    \rowcolor{hlrowcolor}
    \linespread{0.5}\noindent
    \small{\scriptsize A woman is making a surprise face.}
    \end{tabular}
    \includegraphics[width=\linewidth]{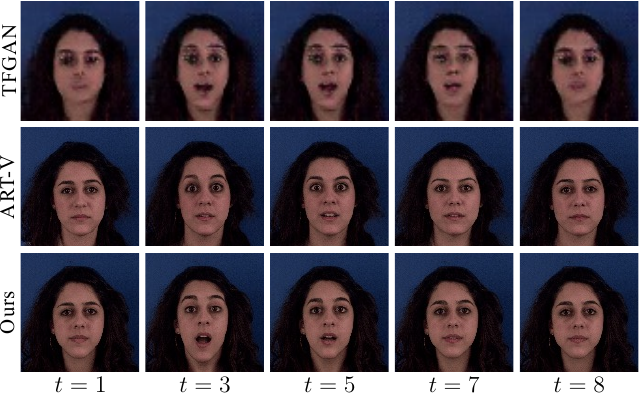}
    \caption{Samples on \textbf{MUG}.
    ART-V and our method can generate sharp and temporally consistent videos while frames 
    produced by TFGAN are blurry.
    }
    \label{fig:mug_txt_all}
\end{subfigure}\hfill
\begin{subfigure}[t] {1\linewidth}
    \centering
    \setlength{\tabcolsep}{0\linewidth}
    \begin{tabular}{L{1\linewidth}}
    \rowcolor{hlrowcolor}
    \linespread{0.5}\noindent
    \scriptsize	A female is wearing lipstick. This person has wavy hair and blond hair. She has high cheekbones, arched eyebrows and rosy cheeks. She has heavy makeup. She is young.
    \end{tabular}
    \includegraphics[width=1\linewidth]{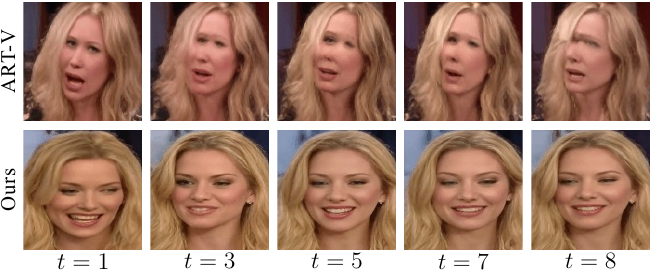}
    \caption{Samples on \textbf{Multimodal VoxCeleb}. Frame generated by ART-V at $t=1$ is sharp and clear, but are blurry at later steps such as $t=5 ~\text{and}~ 8$.
    }
    \label{fig:vox_txt_all}
\end{subfigure}
\caption{\textbf{Text-to-video} generation results for different methods. Sample frames are shown at several time steps ($t$). Conditioned text is provided at the top with light blue background.}
\label{fig:text-to-video-all}
\end{figure}
\begin{table}[h]
\caption{Classification accuracy (\%) on the Shapes dataset for video generation. Our method achieves the best performance.}
    \label{tab:shape_text}
    \centering
    \resizebox{1\linewidth}{!}{
    \begin{tabular}{llcccccc}
    \toprule
    Condition & Methods & Shape & Color & Size & Motion & Dir & Avg \\
    \hline
    \multirow{3}{*}{Text Only} & TFGAN~\cite{tfgan} & 80.22 & \textbf{100.00} & 84.33 & \textbf{99.90} & \textbf{99.95} & 92.88 \\
    & ART-V & 95.07 & 98.68 & 97.71 & 92.72 & 96.04 & 96.04 \\ 
    &\cellcolor{hlrowcolor2}MMVID (Ours) &\cellcolor{hlrowcolor2}\textbf{95.56} &\cellcolor{hlrowcolor2}99.71 &\cellcolor{hlrowcolor2}\textbf{97.95} &\cellcolor{hlrowcolor2}97.80 &\cellcolor{hlrowcolor2}99.61 &\cellcolor{hlrowcolor2}\textbf{98.12} \\
    \hline\hline
    \multirow{2}{*}{Multimodal}
    & ART-V & 92.82 & 97.17 & 97.31 & 89.55 & 93.99 & 94.17 \\
    &\cellcolor{hlrowcolor2}MMVID (Ours) &\cellcolor{hlrowcolor2}\textbf{98.19} &\cellcolor{hlrowcolor2}\textbf{99.76} &\cellcolor{hlrowcolor2}\textbf{98.83} &\cellcolor{hlrowcolor2}\textbf{99.46} &\cellcolor{hlrowcolor2}\textbf{99.95} &\cellcolor{hlrowcolor2}\textbf{99.24} \\
    \bottomrule
    \end{tabular}}
\end{table}
\subsection{Text-to-Video Generation}
\noindent\textbf{Shapes}.
We report the classification accuracy in Tab.~\ref{tab:shape_text} (top four rows) for the Shapes dataset. ART-V and MMVID are trained for $100$K iterations.
Compared with TFGAN~\cite{tfgan}, our model achieves significantly higher classification accuracy for Shape, Size, and Average (Avg) categories. Compared with ART-V, we perform better in all the categories.
Note that our method has slightly lower accuracy on Color, Motion, and Direction (Dir) than TFGAN. The reason might be the VQGAN introduces errors in reconstruction, since the background is diverse and moving, and  it needs to encode these small translations by only $4\times4$ codes.
Note that to have a fair comparison, we do not apply text augmentation when performing comparison with other works.

\begin{table}[h]
\caption{Inception Score (IS) and classification accuracy (\%) on MUG for video generation. We mark `*' to IS values reported in TiVGAN. Our model achieves highest accuracy and IS.}
    \label{tab:mug_text}
    \centering
    \resizebox{1\linewidth}{!}{
    \begin{tabular}{llccc}
    \toprule
   Condition & Methods & Gender (\%) $\uparrow$ & Expression (\%) $\uparrow$ & IS $\uparrow$ \\
    \hline
    \multirow{7}{*}{Text Only}&TGAN~\cite{saito2017temporal}    & - & - & *4.63 \\
    &MoCoGAN~\cite{tulyakov2017mocogan} & - & - & *4.92 \\
     &TGANs-C~\cite{pan2017create} & - & - & *4.65 \\ 
     &TiVGAN~\cite{kim2020tivgan}  & - & - & *5.34 \\ 
     &TFGAN~\cite{tfgan} & 99.22 & \textbf{100.00} & 5.53 \\ 
   &ART-V & 93.46 & 99.12 & 5.72 \\ 
    &\cellcolor{hlrowcolor2}{MMVID} (Ours) &\cellcolor{hlrowcolor2}\textbf{99.90} &\cellcolor{hlrowcolor2}\textbf{100.00} &\cellcolor{hlrowcolor2}\textbf{5.94} \\ 
    \hline\hline
     \multirow{2}{*}{Multimodal} & ART-V & 89.16 & 98.54 & 5.59 \\
    &\cellcolor{hlrowcolor2}MMVID (Ours) &\cellcolor{hlrowcolor2}\textbf{98.14} &\cellcolor{hlrowcolor2}\textbf{100.00} &\cellcolor{hlrowcolor2}\textbf{5.85} \\ 
    \bottomrule
    \end{tabular}}
\end{table}
\begin{table}[h]
\caption{Evaluation metrics for text-to-video generation on iPER and Multimodal VoxCeleb datasets.}
    \label{tab:vox_and_iper}
    \centering
    \resizebox{1\linewidth}{!}{
    \begin{tabular}{llcccc}
    \toprule
    Dataset & Methods & CLIP $\uparrow$ & FVD $\downarrow$ & $F_8$ $\uparrow$ & $F_{1/8}$ $\uparrow$  \\
    \hline
    \multirow{2}{*}{iPER} & ART-V & - & 277.604 & 0.936 & 0.806  \\
    &\cellcolor{hlrowcolor2}MMVID (Ours) &\cellcolor{hlrowcolor2}- &\cellcolor{hlrowcolor2}\textbf{209.127} &\cellcolor{hlrowcolor2}\textbf{0.944} &\cellcolor{hlrowcolor2}\textbf{0.924}  \\ \hline\hline
    \multirow{1}{*}{Multimodal} & ART-V & 0.193 & 60.342 & 0.953 & 0.960  \\
    \multirow{1}{*}{VoxCeleb} &\cellcolor{hlrowcolor2}MMVID (Ours) &\cellcolor{hlrowcolor2}\textbf{0.197} &\cellcolor{hlrowcolor2}\textbf{46.763} &\cellcolor{hlrowcolor2}\textbf{0.972} &\cellcolor{hlrowcolor2}\textbf{0.971}  \\
    \bottomrule
    \end{tabular}}
\end{table}

\noindent\textbf{MUG}.
We follow the experimental setup in TiVGAN~\cite{kim2020tivgan} for experiments on the MUG expression dataset. 
We train models with a temporal step size of $8$ due to the memory limit of GPU. Note TiVGAN is trained with a step size of $4$ and generates $16$-frame videos, while our model generates $8$-frame videos in a single forward. We also train a 3D ConvNet as described in TiVGAN to evaluate the Inception Score and perform classification on Gender and Expression. Results are shown in Fig.~\ref{fig:mug_txt_all} and Tab.~\ref{tab:mug_text} (top 8 rows). Our model achieves the best performance.

\begin{figure*}[t]
    \centering
    \includegraphics[width=1\linewidth]{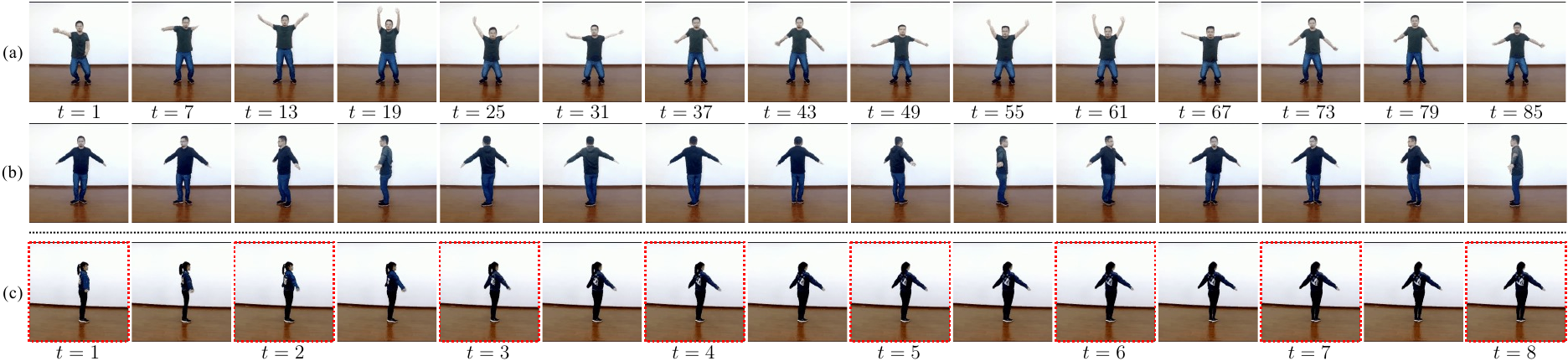}
    \caption{\textbf{Extrapolation and Interpolation}. \emph{Row (a) - (b)}: long sequence generation via extrapolation. 
    \emph{Row (c)}: interpolating a real sequence. Frames in dotted red boxes are fixed as preservation control. Textual controls for each row are: (a) ``Person 024 dressed in 2 is performing random pose, normal speed.''; (b) ``Person 024 dressed in 1 is performing A-pose, normal speed.''; and (c) ``Person 028 dressed in 2 is performing A-pose, normal speed.''}
    \label{fig:iper}
\end{figure*}

\noindent\textbf{iPER}. 
We show the results of the dataset in Tab.~\ref{tab:vox_and_iper} (top 3 rows), demonstrating the advantages of our method. Long sequence generation results are shown in Fig.~\ref{fig:iper}.

\noindent\textbf{Multimodal VoxCeleb}. We train ART-V and our model at a spatial resolution of $128\times128$ and a temporal step of $4$ to generate $8$ frames. Our method shows better results than ART-V on all the metrics, as shown in Tab.~\ref{tab:vox_and_iper} (bottom two rows). We notice ART-V can also generate video samples with good visual quality and are aligned well with the text descriptions. However, ART-V often produces samples that are not temporally consistent. For example, as shown in Fig.~\ref{fig:vox_txt_all}, the frame generated by ART-V at $t=1$ is sharp and clear, but frames at $t=5$ or $t=8$ are blurry. The reason might be the exposure bias in autoregressive models becomes more obvious as the sequence length is long, \emph{i.e.}, an $8$-frame video at resolution $128\times128$ has $512$ tokens. Thanks to bidirectional information during training and inference, our MMVID is able to produce temporally consistent videos. 
\begin{figure}[t]
\centering
\begin{subfigure}[t] {1\linewidth}
    \centering
    \setlength{\tabcolsep}{0\linewidth}
    \begin{tabular}{L{1\linewidth}}
    \rowcolor{hlrowcolor}\linespread{0.5}\noindent
    \small{\scriptsize	 A large green square is moving in a diagonal path in the northeast direction.} 
    \end{tabular}
    \includegraphics[width=1\linewidth]{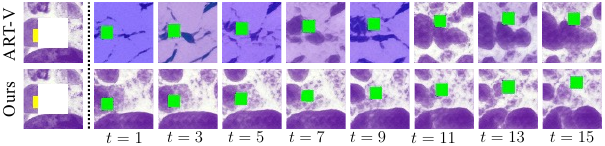}
    \caption{\textbf{Independent} multimodal control.
    The text description specifies the size, color, and shape of the object, and its motion. The visual control is a \emph{partially} observed image with its center masked out (shown as white), which provides control for the background. ART-V can generate correct object and motion, but suffers from incorporating consistent visual inputs such that the background is not temporal consistent. }
    \label{fig:shape_vc}
\end{subfigure}\hfill
\begin{subfigure}[t] {1\linewidth}
    \centering
    \setlength{\tabcolsep}{0\linewidth}
    \begin{tabular}{L{1\linewidth}}
    \rowcolor{hlrowcolor}
    \linespread{0.5}\noindent
    \small{\scriptsize	An object with color in image one, shape in image two, background in image three is moving in a diagonal path in the southwest direction.}
    \end{tabular}
    \includegraphics[width=1\linewidth]{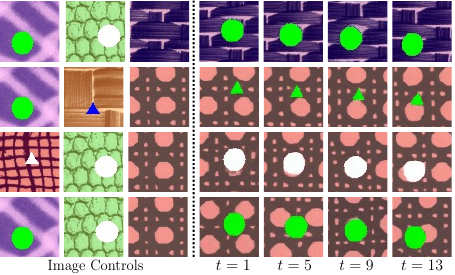}
    \caption{\textbf{Dependent} multimodal controls. The text description specifies from which image to extract color, shape, and background.}
    \label{fig:shape_text_dep}
\end{subfigure}
\caption{\textbf{Multimodal} generation results on Shapes with \emph{textual} (at top) and \emph{visual} ({first} column(s))  modalities. Sample frames are shown at several time step ($t$).}
\label{fig:multimodal-all}
\end{figure}
\begin{figure*}[h]
    \centering
    \includegraphics[width=0.85\linewidth]{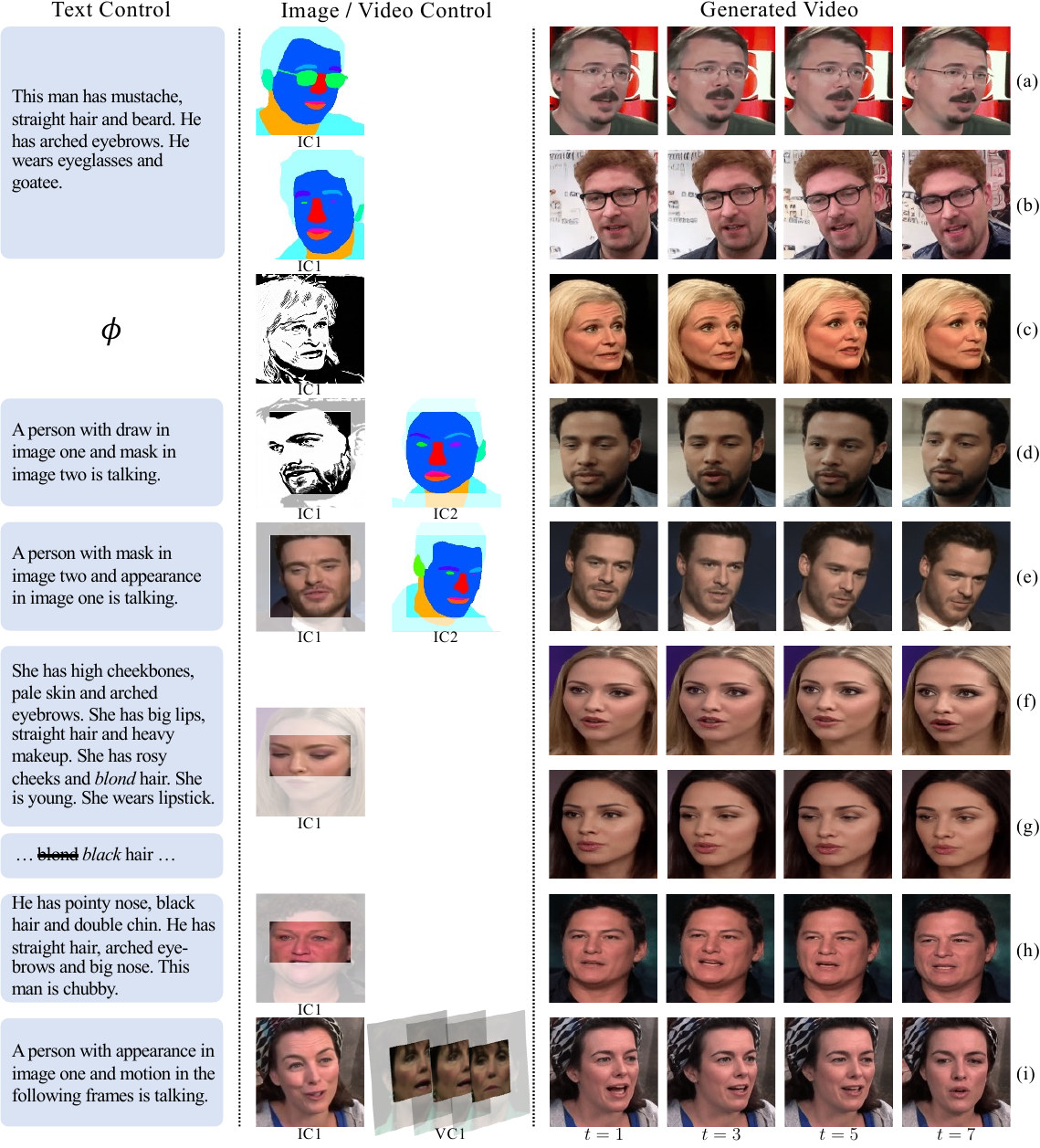}
    \caption{\textbf{Independent and Dependent} multimodel video generation on Multimodal VoxCeleb with textual control (TC), image control (IC), and video control (VC). \emph{Row (a) - (b)}: TC + IC (segmentation mask); \emph{Row (c)}: TC (null) + IC (drawing); \emph{Row (d) - (e)}: dependent TC + IC; \emph{Row (f) - (h)}: TC + IC (partial image) and the TC of (g) is obtained from the TC of (f) by replacing ``blond'' with ``black''; \emph{Row (i)}: dependent TC + VC and the VC includes content and motion information. }
    \label{fig:vox_vc}
\end{figure*}

\subsection{Multimodal Video Generation}
\noindent Multimodal conditions can evolved in two cases: independent and dependent, and we show experiments on both.

\noindent\textbf{Independent Multimodal  Controls}. This setting is similar to conventional conditional video generation, except the condition is changed to multimodal controls.
We conduct experiments on Shapes and MUG datasets with the input condition as the combination of text and image. The bottom two rows in Tab.~\ref{tab:shape_text} and Tab.~\ref{tab:mug_text} demonstrate the advantages of our method over ART-V on all metrics. Additionally, we provide generated samples in Fig.~\ref{fig:shape_vc}, where only a \emph{partial} image is given as the visual condition. As can be seen, ART-V cannot satisfy the visual constraint well and the generated video is not consistent. 
The quality degradation for multimodal video synthesis of ART-V can also be verified in Tab.~\ref{tab:shape_text} as it shows lower classification accuracy than text-only generation, while our method is able to generate high quality videos for different condition signals.

We also conduct extensive experiments of video generation under various combinations of textual and image controls on Multimodal VoxCeleb, as shown in Fig.~\ref{fig:vox_vc}. We apply three different image controls, including segmentation mask (Fig.~\ref{fig:vox_vc} row (a) - (b)), drawing (Fig.~\ref{fig:vox_vc} row (c) - (d)), and partial image (Fig.~\ref{fig:vox_vc} row (d) - (f)).  In Fig.~\ref{fig:vox_vc} row (b), our method can synthesize frames with eyeglasses even though eyeglasses are not shown in segmentation mask. In  Fig.~\ref{fig:vox_vc} row (f) - (g), we show that using the same image control while  replacing the ``blond'' with ``black'' in the text description, we can generate frames with similar content except the hair color is changed. Such examples demonstrate that our method has a good understanding of multimodal controls.

\noindent\textbf{Dependent Multimodal Controls}. Furthermore, we introduce a novel task for multimodal video generation where textual and visual controls are dependent, such that the actual control signals are guided by the textual description. For example, Fig.~\ref{fig:shape_text_dep} illustrates how the text informs from which image  the model should query color, shape, and background information. More synthesized examples on Multimodal VoxCeleb are given in Fig.~\ref{fig:vox_vc}. For Fig.~\ref{fig:vox_vc} row (d) - (e), 
our model learns to combine detailed facial features from drawing or image and coarse features (\emph{i.e.}, pose) from mask.
For Fig.~\ref{fig:vox_vc} row (i), our method successfully retargets the subject with an appearance from the given image control (IC1) and generates frames with the motion specified by consecutive images that provide motion control (VC1).

\begin{table}[h]
    \caption{Analysis on Shapes for video augmentation strategies.
    }
    \label{tab:ablation_vid}
    \centering
    \resizebox{1\linewidth}{!}{
    \begin{tabular}{cccccccccc}
    \toprule
    \multicolumn{4}{c}{Video Augmentation} & \multicolumn{6}{c}{Accuracy (\%)}\\
    \cmidrule(lr){1-4} \cmidrule(lr){5-10}
    Swap & Shuffle & Color & Affine & Shape & Color & Size & Motion & Dir & Avg \\
    \hline
    \redxmark & \redxmark & \redxmark & \redxmark & 90.43 & 89.07 & 95.61 & 92.48 & 99.13 & 93.34 \\
    \greencmark & \redxmark & \redxmark & \redxmark & 91.02 & 89.84 & 93.75 & 91.02 & 98.05 & 92.73 \\
    \redxmark & \greencmark & \redxmark & \redxmark & 88.28 & 89.45 & 94.53 & 88.28 & 98.44 & 91.80 \\
    \redxmark & \redxmark & \greencmark & \redxmark & 91.80 & {\bf 91.02} & 94.53 & 93.36 & 98.83 & 93.91 \\
    \redxmark & \redxmark & \redxmark & \greencmark & 90.62 & 90.62 & 95.31 & 89.84 & 98.83 & 93.05 \\
    \rowcolor{hlrowcolor2}
    \greencmark & \greencmark & \greencmark & \greencmark & {\bf 93.36} & 88.28 & {\bf 95.70} & {\bf 93.75} & {\bf 99.61} & {\bf 94.14} \\
    \bottomrule
    \end{tabular}}

\end{table}
\subsection{Long Sequence Generation and Ablation}
\noindent\textbf{Long Sequence Generation}.
Our approach enables temporal extrapolation of videos. We show samples of video extrapolation and interpolation in Fig.~\ref{fig:iper}. Samples from Fig.~\ref{fig:iper} row (a) - (b) are generated by being iteratively conditioned on previous $6$ frames to generate the following $2$ frames. Fig.~\ref{fig:iper} row (c) shows an example of synthesizing one frame by interpolating two consecutive real frames.

\begin{figure}[h]
    \centering
    \includegraphics[width=0.95\linewidth]{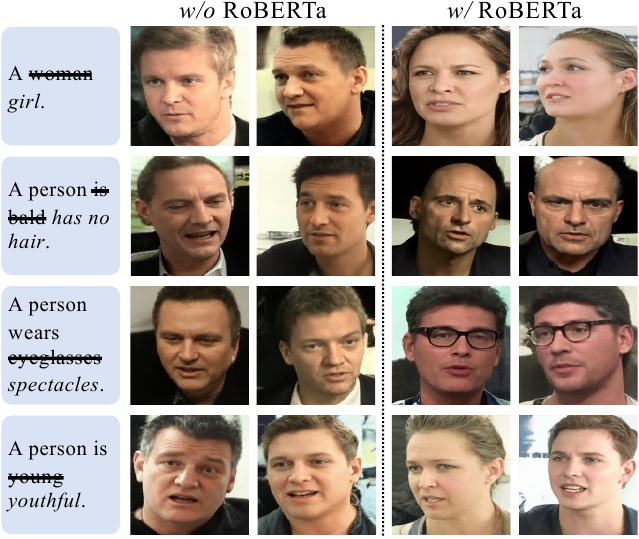}
    \caption{\textbf{Analysis on language embedding}. Samples are generated with out-of-distribution textual inputs.
    We reword the original text (strikethrough) with equivalent descriptions (italic) that do not exist in the training.
    We show the first frames from the generated sequences for each method. Frames generated using the pretrained language model (\emph{w/} RoBERTa) is more correlated with text inputs.}
    \label{fig:vox_lang}
\end{figure}
\noindent\textbf{Analysis on VID Task}. We perform analysis for different \texttt{VID} strategies on the Shapes dataset. Tab.~\ref{tab:ablation_vid} shows that the highest average accuracy is achieved when all augmentation is used (sampled uniformly).
Also note that accuracy for color is the highest when we only apply color augmentation. 

\noindent\textbf{Analysis on Language Embedding}.  Analysis of using a pretrained language model is shown in Fig.~\ref{fig:vox_lang}. The method with a language model (\emph{w/} RoBERTa) is more robust to various text inputs than the one without it (\emph{w/o} RoBERTa).

\section{Limitation and Conclusion}
This paper targets a new problem, which is video generation using multimodal inputs. To tackle the problem, we utilize a two-stage video generation framework that includes an autoencoder for quantized representation of images and videos and a non-autoregressive transformer for predicting video tokens from multimodal input signals. Several techniques are proposed, including the special \texttt{VID} token, textual embedding, and improved mask prediction, to help generate temporally consistent videos. 
On the other hand, the proposed method also contains some limitations, including temporal consistency issues for high-resolution videos, generating diverse motion patterns for longer sequences, and further improving the diversity of non-autoregressive transformers. More details can be found in the Appendix. Besides improving the limitation, a future direction might be to leverage more control modalities, such as audio, to generate videos with a much higher resolution. 

{\small
\bibliographystyle{ieee_fullname}
\bibliography{ref}
}
\clearpage
\appendix
\section*{Appendix}
\addcontentsline{toc}{section}{Appendix}
{\hypersetup{linkcolor=blue}
\startcontents[sections]
\printcontents[sections]{l}{1}{\setcounter{tocdepth}{2}}
}

\section{More Details and Ablation for Methods}
In this section, we introduce additional details of our methods. Specifically, we describe the settings for the masking strategies for Masked Sequence Modeling (MSM) in Sec.~\ref{sec:msm}, different training methods for Relevance Estimation (REL) task in Sec.~\ref{sec:rel}, augmentation performed on the task of Video consistency modeling (VID)
in Sec.~\ref{sec:vid}, additional discussion on improved mask-predict for video prediction in Sec.~\ref{sec:mask_predict}, and an ablation analysis on text augmentation in Sec.~\ref{sec:text_aug}.

\subsection{Settings for Masking Strategies in MSM Task}\label{sec:msm}

In the main paper (Sec. 3.1), we introduce five masking strategies, \emph{i.e.}, (I) i.i.d. masking;
(II) masking all tokens; (III) block masking; (IV) the negation of block masking; and (V) randomly keeping some frames, to train the task of mask sequence modeling. In all of our experiments, if not specified, we apply strategies I - IV with probabilities as $[0.7, 0.1, 0.1, 0.1]$.
For strategy V, we adopt it by randomly keeping $k$ frames on top of the mask produced from strategies I - IV.
We set the probability of strategy V as $0.2$ and $k=T/2$, where $T$ is the total number of frames.

\subsection{Training Methods for REL Task}\label{sec:rel}
We compare two training methods for the relevance estimation task. The first one is swapping the conditional inputs to get the negative sample, which we denote as \texttt{REL}=swap. The method is introduce in Sec. 3.1 of the main paper. The second method, \texttt{REL}=negative, is to sample a negative training data such that it has a different annotation as the positive one. This ensures that the negative sequence for REL is indeed negative, which is not guaranteed in the case of conditional swapping. As shown in Tab.~\ref{tab:ablation_rel}, we empirically find that negative sampling achieves better performance than conditional swapping in the early stage but its FVD and $F_8$ score becomes inferior when the model converges. Thus, \texttt{REL}=swap is used in all experiments if not specified.

\begin{table}[h]
\caption{Analysis of the use of different training methods for the video relevance task on the Multimodal VoxCeleb dataset. Results from two iterations ($50$K and $100$K) are reported for each method.}
    \label{tab:ablation_rel}
    \centering
    \resizebox{1\linewidth}{!}{
    \begin{tabular}{lllccc}
    \toprule
    Resolution & Method & Iteration & FVD $\downarrow$ & $F_8$ $\uparrow$ & $F_{1/8}$ $\uparrow$  \\
    \hline
    \multirow{4}{*}{$128\times128$} & \multirow{2}{*}{\texttt{REL}=swap} & 50K & 123.147 & 0.921 & 0.888 \\
    & &\cellcolor{hlrowcolor2}100K &\cellcolor{hlrowcolor2}{\bf 103.622} &\cellcolor{hlrowcolor2}{\bf 0.936} &\cellcolor{hlrowcolor2}{0.895} \\
    \cmidrule(lr){3-6}
    & \multirow{2}{*}{\texttt{REL}=negative} & 50K & { 109.471} & { 0.931} & { 0.903} \\
    & & 100K & 117.128 & 0.922 & {\bf 0.922} \\
    \hline\hline
    \multirow{4}{*}{$256\times256$} & \multirow{2}{*}{\texttt{REL}=swap} & 50K & 293.999 & 0.753 & 0.692 \\
    & &\cellcolor{hlrowcolor2}100K &\cellcolor{hlrowcolor2}{\bf 191.910} &\cellcolor{hlrowcolor2}{\bf 0.781} & \cellcolor{hlrowcolor2}{0.788} \\
    \cmidrule(lr){3-6}
    & \multirow{2}{*}{\texttt{REL}=negative} & 50K & 225.043 & 0.648 & 0.651 \\
    & & 100K & 201.702 & 0.774 & {\bf 0.864} \\
    \bottomrule
    \end{tabular}}
\end{table}

\subsection{Video Augmentation on VID Task}\label{sec:vid}
\begin{figure}[h]
    \centering
    \includegraphics[width=0.9\linewidth]{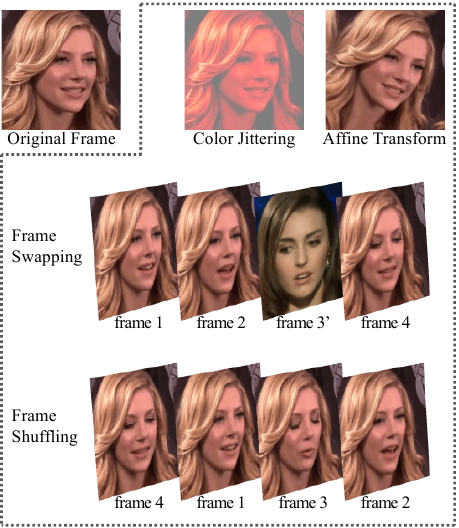}
    \caption{\textbf{Augmentation strategies for modeling video consistency.} \emph{Top Row}: first column -- original frame; second column -- augmented with color jittering; third column -- augmented with affine transform.
    \emph{Second Row}: frame swapping such that the third frame is swapped by using a frame from another video.
    \emph{Third Row}: frame shuffling such that the position of frames is randomly shuffled.
    }
    \label{fig:vid_aug}
\end{figure}

We propose a \texttt{VID} token to for modeling video consistency (Sec. 3.2 in the main paper). To learn the \texttt{VID} in a self-supervised way, we introduce four negative video augmentation methods. Here we illustrate more details for each augmentation strategy, shown in Fig.~\ref{fig:vid_aug}, including  color jittering, affine transform, frame swapping, and frame shuffling.
In all of our experiments, if not specified, we uniformly sample these strategies with probabilities as $[0.25, 0.25, 0.25, 0.25]$.

\subsection{More Details on Improved Mask-Predict}\label{sec:mask_predict}
\noindent\textbf{\texttt{SampleToken} and \texttt{SampleMask}}. 
We introduce our  algorithm for improved mask-predict in the main paper (Alg. 1). Here we provide more details of the two functions (\texttt{SampleToken} and \texttt{SampleMask}) used in the algorithm. 

\begin{itemize}[leftmargin=1em]
  \setlength\itemsep{-0.25em}
    \item \texttt{SampleToken} is given in Alg.~\ref{alg:sample_token}, with PyTorch~\cite{NEURIPS2019_9015}-like functions. $\texttt{Gather}(\mathbf{p}, \mathbf{z})$ gathers values of $\mathbf{p}$,
    which is a matrix whose dimensions are the number of tokens by the number of words,
    along the token axis specified by indices $\mathbf{z}$.
    \item \texttt{SampleMask} is given in Alg.~\ref{alg:sample_mask}. The function $\texttt{Find}$ collects the indices of the $\texttt{True}$ elements; function $\texttt{Multinomial}(\mathbf{y}, n)$ samples $n$ points without replacement from a multinomial specified by $\mathbf{y}$ and returns their indices; and function $\texttt{Scatter}(\mathbf{0}, \mathbf{j}, 1)$ sets values to $1$ in a tensor initialized to $\mathbf{0}$ at locations specified by indices $\mathbf{j}$. Lines 1 - 5 in Alg.~\ref{alg:sample_mask} sample $n$ locations, according to $\mathbf{y}$, to be preserved, and the locations with $\mathbf{m}_\text{PC}$ equal to $1$  are always selected.
\end{itemize}

\begin{algorithm}[h]
\caption{\texttt{SampleToken}}
\label{alg:sample_token}
\begin{algorithmic}[1]
    \Require Logit $\tilde{\mathbf{p}}$ and noise level $\sigma$.
    \State $\mathbf{g} \leftarrow \texttt{Gumbel}(0, 1) ~~i.i.d.$
    \State $\mathbf{p} \leftarrow \texttt{Softmax}(\tilde{\mathbf{p}}+\sigma\mathbf{g})$
    \State $\mathbf{z} \leftarrow \texttt{Multinomial}(\mathbf{p})$ \Comment{\textcolor{commentcolor}{sample from multinomial}}
    \State $\mathbf{y} \leftarrow \texttt{Gather}(\mathbf{p}, \mathbf{z})$ \Comment{\textcolor{commentcolor}{collect probs for each token}}
    \State \Return $\mathbf{z}$, $\mathbf{y}$
\end{algorithmic}
\end{algorithm}
\begin{algorithm}[h]
\caption{\texttt{SampleMask}}
\label{alg:sample_mask}
\begin{algorithmic}[1]
    \Require Probabilities $\mathbf{y}$, preservation mask $\mathbf{m}_\text{PC}$, and the number of tokens to keep $n$.
    \State $\mathbf{y}' \leftarrow \mathbf{y}\texttt{[}\mathbf{m}==0\texttt{]}$ \Comment{\textcolor{commentcolor}{collect probs no need to preserve}}
    \State $\mathbf{i}' \leftarrow \texttt{Find}(\mathbf{m}==0)$ \Comment{\textcolor{commentcolor}{collect indices}}
    \State $\mathbf{i} \leftarrow \texttt{Multinomial}(\text{Normalize}(\mathbf{y}'), n)$
    \State $\mathbf{j} \leftarrow \mathbf{i}'\texttt{[}\mathbf{i}\texttt{]}$ \Comment{\textcolor{commentcolor}{slicing to get sampled indices}}
    \State $\mathbf{m} \leftarrow \texttt{Scatter}(\mathbf{0}, \mathbf{j}, 1)$ \Comment{\textcolor{commentcolor}{populate indices}}
    \State $\mathbf{m} \leftarrow \mathbf{m} \texttt{|} \mathbf{m}_\text{PC}$ \Comment{\textcolor{commentcolor}{elementwise OR}}
    \State \Return $\mathbf{m}$
\end{algorithmic}
\end{algorithm}

\begin{figure*}[h]
    \centering
    \includegraphics[width=1\linewidth]{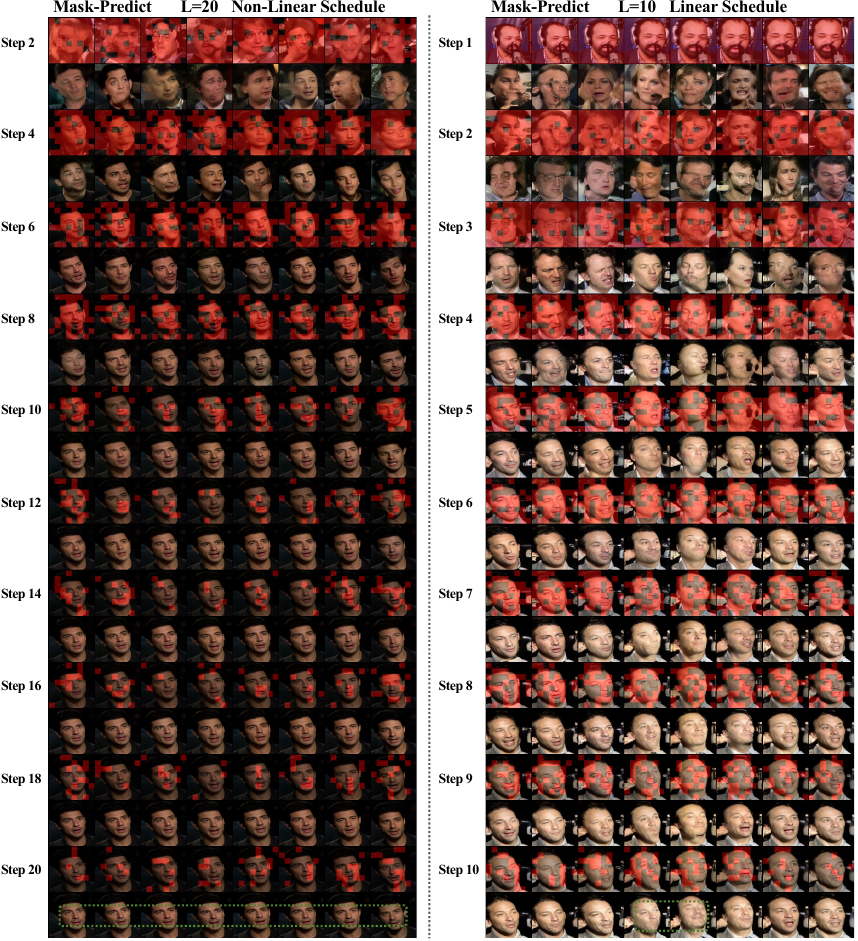}
    \caption{\textbf{Comparison between non-linear (ours) \emph{vs.} linear schedule for mask annealing in mask-predict}. 
    The mask-predict starts from a fully-masked sequence (Step $1$, and the images displayed beneath red masks in Step $1$ are real video frames). Patches with red color denote the corresponded tokens are masked. The images with red color are generated after the mask-predict at that step. 
    \emph{Left}: frames generated using our non-linear mask annealing scheme. The motion is vivid and frames have high quality (highlighted in dotted green box). \emph{Right}: using a linear scheme ($L=L_1=10, \alpha_1=0.9, \beta_1=0.1$) to generate frames. Artifacts can be observed on the synthesized images (highlighted in dotted green box). Frames are synthesized from the model trained on the Multimodal VoxCeleb dataset.
    }
    \label{fig:mp1}
\end{figure*}
\begin{figure*}[h]
    \centering
    \includegraphics[width=1\linewidth]{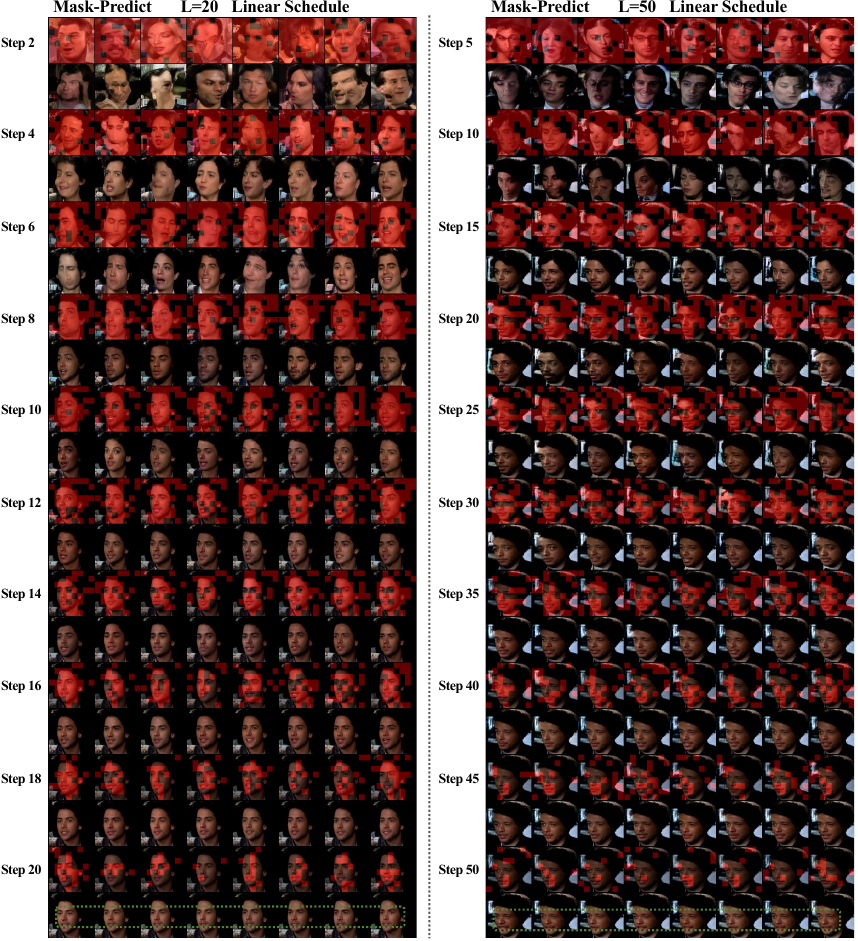}
    \caption{
    \textbf{Frames synthesized by using linear schedule for mask annealing in mask-predict}. 
    The mask-predict starts from a fully-masked sequence (Step $1$, and the images displayed beneath red masks in Step $1$ are real video frames). Patches with red color denote the corresponded tokens are masked. The images with red color are generated after the mask-predict at that step. Two samples have the setting as $\alpha_1=0.9$ and $\beta_1=0.1$. Motion has been washed out, \emph{i.e.}, frames in a sequence tends to be static and have similar appearance as illustrated in dotted green box, 
    for the setting of $L=L_1=20$ (\emph{Left}) and $L=L_1=50$ (\emph{Right}). Frames are synthesized from the model trained on the Multimodal VoxCeleb dataset.
    }
    \label{fig:mp2}
\end{figure*}

\begin{figure*}[h]
\centering
\begin{subfigure}[t] {0.48\linewidth}
    \centering
    \setlength{\tabcolsep}{0\linewidth}
    \begin{tabular}{L{1\linewidth}}
    \rowcolor{hlrowcolor}\linespread{0.5}\noindent
    \small{\scriptsize	 She has rosy cheeks, blond hair and arched eyebrows. She wears lipstick and earrings. She is young.} 
    \end{tabular}
    \includegraphics[width=0.9\linewidth]{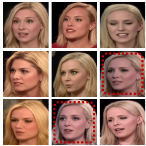}
    \caption{\emph{W/} noise annealing in mask-predict.
    }
    \label{fig:mp_addnoise}
\end{subfigure}\hfill
\begin{subfigure}[t] {0.48\linewidth}
    \centering
    \setlength{\tabcolsep}{0\linewidth}
    \begin{tabular}{L{1\linewidth}}
    \rowcolor{hlrowcolor}
    \linespread{0.5}\noindent
    \small{\scriptsize	She has rosy cheeks, blond hair and arched eyebrows. She wears lipstick and earrings. She is young.}
    \end{tabular}
    \includegraphics[width=0.9\linewidth]{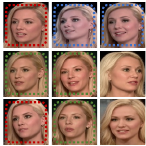}
    \caption{\emph{W/O} noise annealing in mask-predict.
    }
    \label{fig:mp_nonoise}
\end{subfigure}
\caption{ \textbf{Comparison between \emph{w/} (a) and \emph{w/o} (b)  noise annealing in mask-predict for text-to-video generation}. Each image in a subfigure is the first frame sampled from a synthesized video and each subfigure includes $9$ videos.
For the two subfigures, we use the same textual input to generate videos and apply dotted boxes with the same color to denote the synthesized videos with the same (or very similar) identity. Adding noise improves diversity as (a) only contains two images with the same (or very similar) identity.
Frames are generated from a model trained on the Multimodal VoxCeleb dataset. We use linear mask annealing scheme with $L=15$.
}
\label{fig:mp_noise}
\end{figure*}

\noindent\textbf{Mask Annealing}.
We define the piecewise linear mask annealing scheme $n^{(i)}$ (used in Alg. 1 in the main paper) as follows. 
\begin{equation}
\resizebox{1\linewidth}{!}{%
    $n^{(i)} = \left\{ \begin{array}{lcl}
         N\cdot (\beta_1+\frac{L_1-i}{L_1-1}\cdot(\alpha_1-\beta_1)) & \mbox{for}
         & 1\leq i \leq L_1 \\ 
          N\cdot \alpha_2  & \mbox{for} & L_1 < i \leq L_1+L_2 \\
         N\cdot \alpha_3  & \mbox{for} & i > L_1+L_2
    \end{array}\right.$
    }\label{eq:mask_annealing}
\end{equation}
where we set $L_1=10$, $L_2=10$, $\alpha_1=0.9$, $\beta_1=0.1$, $\alpha_2=0.125$, and $\alpha_3=0.0625$.
We use the following values in experiments if not specified: $L_1=10, L_2=10, \alpha_1=0.9, \beta_1=0.1, \alpha_2=0.125, \alpha_3=0.0625$. The total step of mask-predict is $L$ so that $L_1 + L_2 \leq L$.

Compared with linear annealing, our non-learning mask annealing can generate videos with vivid motion and less artifacts. Example samples for models trained on the Multimodal VoxCeleb dataset are illustrated in Fig.~\ref{fig:mp1} and Fig.~\ref{fig:mp2}. Our method generates facial videos with high fidelity for $L$ as $20$ (Fig.~\ref{fig:mp1}, left), while linear annealing generates low quality frames (Fig.~\ref{fig:mp1}, right) and static videos where motion can hardly be observed (Fig.~\ref{fig:mp2}).

\noindent\textbf{Noise Annealing}.
We define the noise annealing schedule $\sigma^{(i)}$ as follows:
\begin{equation}
\resizebox{1\linewidth}{!}{%
    $\sigma^{(i)} = \left\{ \begin{array}{lcl}
         \beta_1+\frac{L_1-i}{L_1-1}\cdot(\alpha_1-\beta_1) & \mbox{for}
         & 1\leq i \leq L_1 \\ 
         \alpha_2  & \mbox{for} & L_1 < i \leq L_1+L_2 \\
         \alpha_3  & \mbox{for} & i > L_1+L_2
    \end{array}\right.$
    }\label{eq:mp_noise}
\end{equation}
where $L_1, \alpha_1, \beta_1$ are reused from Eqn.~\ref{eq:mask_annealing} for simplicity of notation, but with different values. We set $L_1=10, L_2=5, \alpha_1=0.4, \beta_1=0.02, \alpha_2=0.01, \alpha_3=0$.

Adding noise improves diversity for generated videos, as shown in Fig.~\ref{fig:mp_noise}. However, there is a tradeoff between diversity and quality.
Adding too much noise influences sample quality, which might be due to unconfident tokens that cannot be remasked.

\noindent\textbf{Beam Search}.
We analysize different numbers of beams $B$ employed in the beam search that is used in mask-predict. Results shown in Tab.~\ref{tab:mp_beam} show that using $B=15$ achieves the best results. Interestingly, we empirically find that increasing $B$ from 15 to 20 causes performance drop. We hypothesize that this is due to the scores used for beam selection is not accurate. When $B$ gets larger, the negative influence of inaccurate score estimation becomes more prominent. We use $B=3$ in experiments if not specified.

\begin{figure*}[h]
    \centering
    \includegraphics[width=1\linewidth]{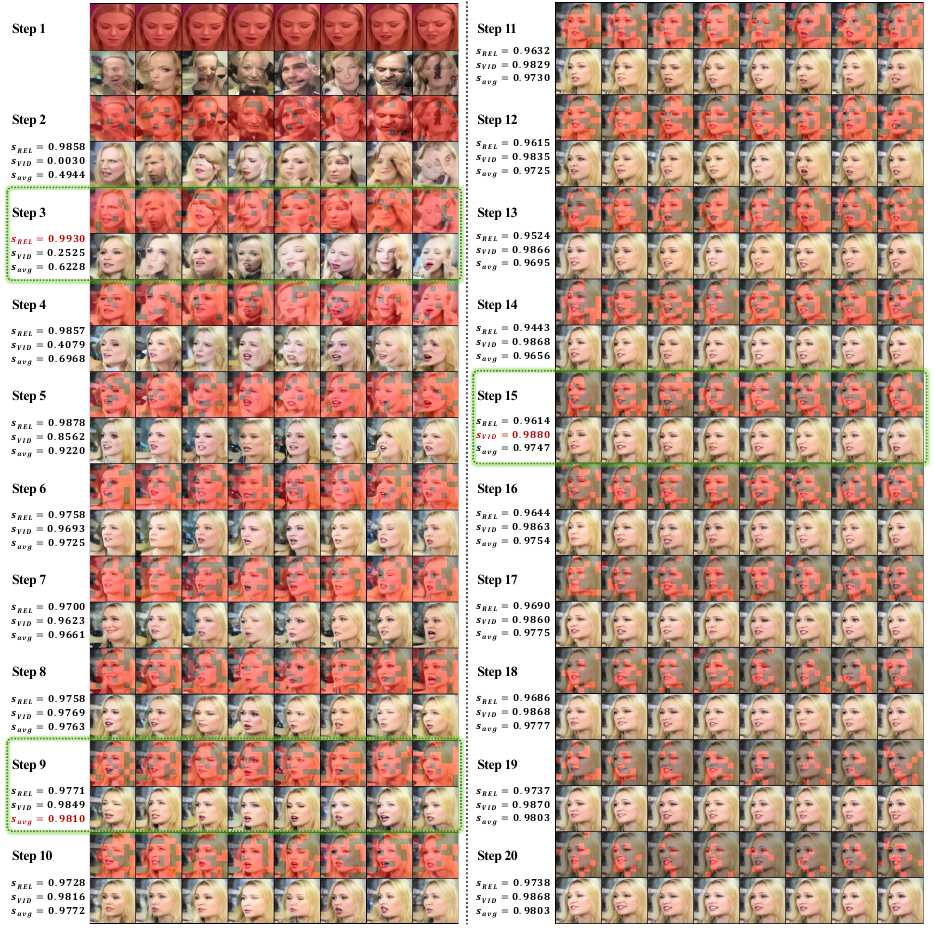}
    \caption{\textbf{Visual samples for analyzing early-stop.} REL score $S_\texttt{REL}$ is very high at the beginning and peaks at step $3$. VID score $S_\texttt{VID}$ peaks at step $15$. The average score $S_\texttt{avg}$ reaches the highest value at step $9$. Step $1$ shows mask-predict starts from a fully-masked sequence and the images displayed beneath red masks in Step 1 are real video frames. $B=1$ is used.}
    \label{fig:mp_early}
\end{figure*}

\noindent\textbf{Early-Stop}.
Early-stop is proposed in previous text-to-image generation~\cite{zhang2021ufc} to stop the mask-predict at the earlier iteration for faster inference. Here we analyze the use of early-stop in our work, and determine that it cannot improve the efficiency. We obtain the scores $S_\texttt{REL}$ and $S_\texttt{VID}$, calculated from two special tokens \texttt{RED} and \texttt{VID}, respectively. We denote $S_\texttt{avg}$ as their averaged score and use $S_\texttt{avg}$ to decide the iteration for stopping if the highest score does not change for $3$ iterations. We first show the quantitative results in Tab.~\ref{tab:mp_beam}, where we find that early-stop does not improve the FVD at $B=1, \ 3, \ 5$. We further provide visual images in Fig.~\ref{fig:mp_early}.
We can see that $S_\texttt{REL}$ is very high at the beginning, and peaks at step $3$, $S_\texttt{VID}$ peaks at step $15$, and the average score $S_\texttt{avg}$ reaches the highest value at step $9$. However, we can still observe artifacts at each step ($3$, $9$, and $15$).
Therefore, using scores calculated from special tokens might not be a reliable signal for determining early-stop, and we thus decide not to use it in our implementation.

\begin{table}[h]
    \centering
    \caption{\textbf{Analysis on Beam Searching and Early-Stop}. Metrics are evaluated on models trained on the Multimodal VoxCeleb dataset, with the different number of beams $B$ and whether early-stop is enabled. The task is text-to-video generation.}
    \label{tab:mp_beam}
    \begin{tabular}{lcccc}
    \toprule
    $B$ & Early-Stop & FVD $\downarrow$ & $F_8$ $\uparrow$ & $F_{1/8}$ $\uparrow$  \\
    \hline
    1  & \redxmark & 97.992 & 0.939 & 0.930 \\
    1  & \greencmark & 97.957 & 0.917 & 0.928 \\
    3  & \redxmark & 96.288 & 0.945 & 0.925 \\
    3  & \greencmark & 99.899 & 0.930 & 0.929 \\
    5  & \redxmark & 94.170 & 0.922 & 0.937 \\
    5  & \greencmark & 97.908 & 0.923 & 0.925 \\
    10 & \redxmark & 95.560 & 0.932 & 0.924 \\
    15 & \redxmark & 92.828 & 0.933 & 0.933 \\
    20 & \redxmark & 97.247 & 0.922 & 0.918 \\
    \bottomrule
    \end{tabular}
    
\end{table}

\begin{table}[h]
  \caption{Human preference evaluation for different methods on the Multimodal VoxCeleb dataset. The task is text-to-video generation.}
    \label{tab:vox_amt}
    \centering
    \small
    \begin{tabular}{rclc}
    \toprule
    \multicolumn{3}{c}{Methods for Pairwise Comparison} & Human Preference \\ \hline
    MMVID &\emph{vs.}& ART-V & \textbf{54.0}\% : 46.0\% \\
    MMVID-TA &\emph{vs.}& MMVID & \textbf{54.5}\% : 45.5\% \\
    \rowcolor{hlrowcolor2}
    MMVID-TA &\emph{vs.}& ART-V & \textbf{61.2}\% : 38.8\% \\
    \bottomrule
    \end{tabular}
\end{table}

\subsection{Analysis on Text Augmentation}\label{sec:text_aug}
Sec. 3.4 of the main paper introduces text augmentation to improve the correlation between the generated videos and input textual controls.  We also notice that text augmentation can help improve the diversity of the synthesized videos. We performance human evaluation using Amazon Mechanical Turk (AMT) to verify the quality and diversity of videos synthesized from various methods. We consider three comparisons, including \emph{MMVID}, which is our baseline model, \emph{MMVID-TA}, which uses text augmentation, and \emph{ART-V}, which uses the autoregressive transformer. $600$ synthesized videos  on the Multimodal VoxCeleb dataset for the text-to-video generation task are presented to AMT, and the results are shown in Tab.~\ref{tab:vox_amt}. Using text augmentation can help improve the quality and diversity of the generated videos, as $61.2$\% users prefer the MMVID-TA over ART-V.

\subsection{Analysis on Conditioning Partially Occluded Images}
We perform experiments Multimodal VoxCeleb dataset to evaluate whether the features from partially occluded images, when given as the condition, are used for generating videos. The experiments are conducted on Amazon Mechanical Turk (AMT) by asking AMT workers to specify if features from the occluded visual modality input appear in the video. We show three settings: text and partially occluded image, segmentation, and drawing. Each setting contains $200$ videos, with half of them having conditions that do not describe the same identity. Each video is analyzed by $10$ workers.
Results shown in Tab.~\ref{tab:paritial_image} suggest $\sim$70\% of videos contain features from occluded visual modalities.

\begin{table}[h]
  \caption{Human preference evaluation for the multimodal video generation with the setting as text and partially occluded image. Videos are generated from models trained on the Multimodal VoxCeleb dataset. }\label{tab:paritial_image}
    \centering
    \begin{tabular}{l c}
    \toprule
    Condition & Preference \\ \midrule
    Text + Partially Occluded Image & 68.95\% \\
    Text + Partially Occluded Segmentation & 69.70\%  \\
    Text + Partially Occluded Drawing & 70.45\%  \\ 
    \bottomrule
    \end{tabular} 
\end{table}

\section{More Experimental Details}
In this section, we introduce more implementation details in experiments and additional experimental results.
\subsection{More Implementation Details}
\noindent\textbf{Training of Autoencoder}.
For each dataset at each resolution, we finetune an autoencoder from VQGAN model~\cite{esser2020taming} pretrained on ImageNet~\cite{russakovsky2015imagenet}, with $f=16$ (which is the equivalent patch-size a single code corresponds to) and $|\mathcal{Z}|=1024$ (which is the vocabulary size of the codebook).

\noindent\textbf{Evaluation Metrics}. To evaluate the model performance on the Shapes dataset, we train a classifier following the instructions of TFGAN~\cite{tfgan} as the original model is not released. To have a fair comparison, we also retrain a TFGAN model for text-to-video generation. To compute the CLIP~\cite{radford2021learning} score for a video sequence, we calculate the CLIP similarity for each frame in a video and use the maximum value as the CLIP score for the video.

\begin{figure}[h]
    \centering
    \includegraphics[width=1\linewidth]{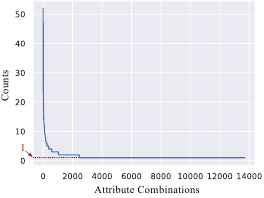}
    \caption{\textbf{Statistics of annotations for the Multimodal VoxCeleb dataset}. The attribute combinations show a long-tail distribution. There are $13,706$ unique attribute combinations out of $19,522$ samples, and $11,259$ combinations have only one data point.}
    \label{fig:vox_stat}
\end{figure}

\begin{figure*}
\begin{minipage}[t]{0.49\linewidth}
\centering
    \includegraphics[width=1\linewidth]{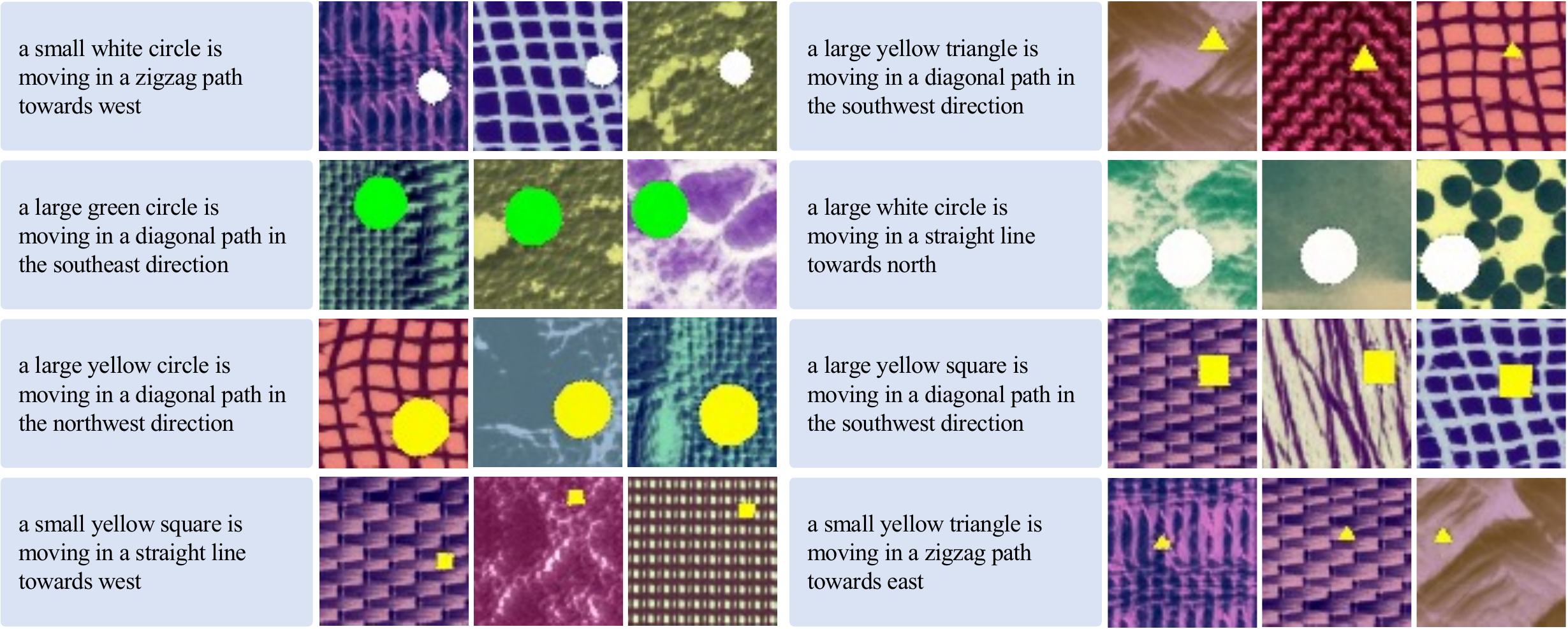}
    \caption{Example videos generated by our approach on the Shapes dataset for text-to-video generation. We show three synthesized videos for each input text condition.}
    \label{fig:more_shape_our_text}
\end{minipage} \hfill
\begin{minipage}[t]{0.49\linewidth}
\centering
   \includegraphics[width=1\linewidth]{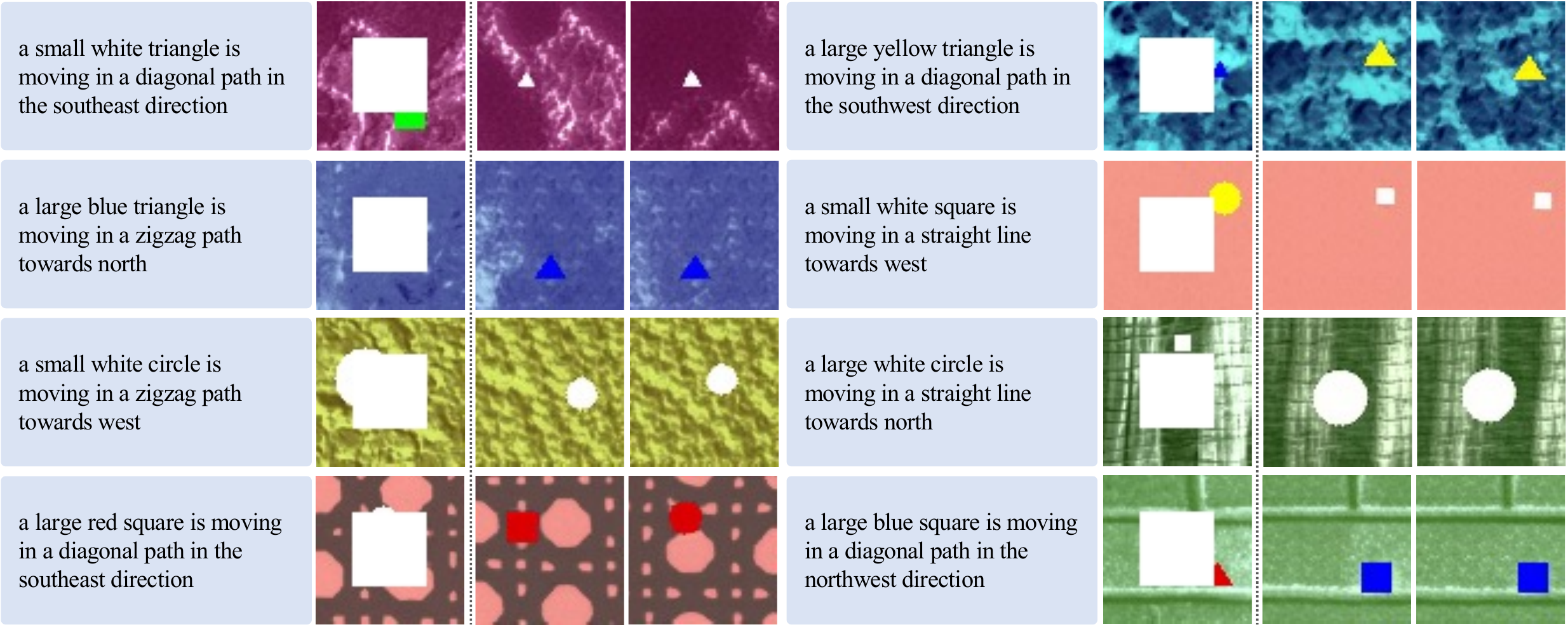}
    \caption{Example videos generated by our approach on the Shapes dataset for independent multimodal generation. The input control signals are text and a partially observed image (with the center masked out, shown in white color). We show two synthesized videos for each input multimodal condition.}
    \label{fig:more_shape_our_text_ic}
\end{minipage} \hfill
\begin{minipage}[t]{0.49\linewidth}
\centering
    \includegraphics[width=1\linewidth]{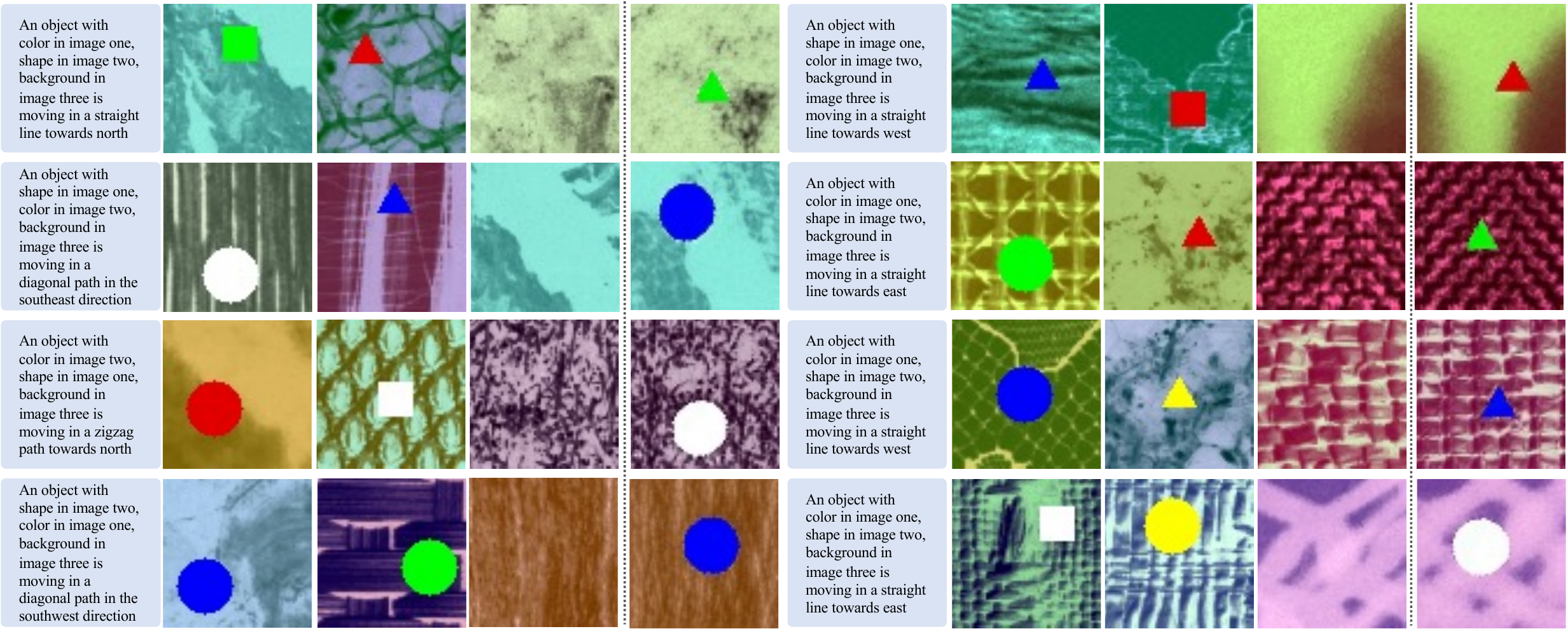}
    \caption{Example videos generated by our approach on the Shapes dataset for dependent multimodal generation. The input control signals are text and images. We show one synthesized video for each input multimodal condition.}
    \label{fig:more_shape_our_depend}
\end{minipage} \hfill
\begin{minipage}[t]{0.49\linewidth}
\centering
    \includegraphics[width=1\linewidth]{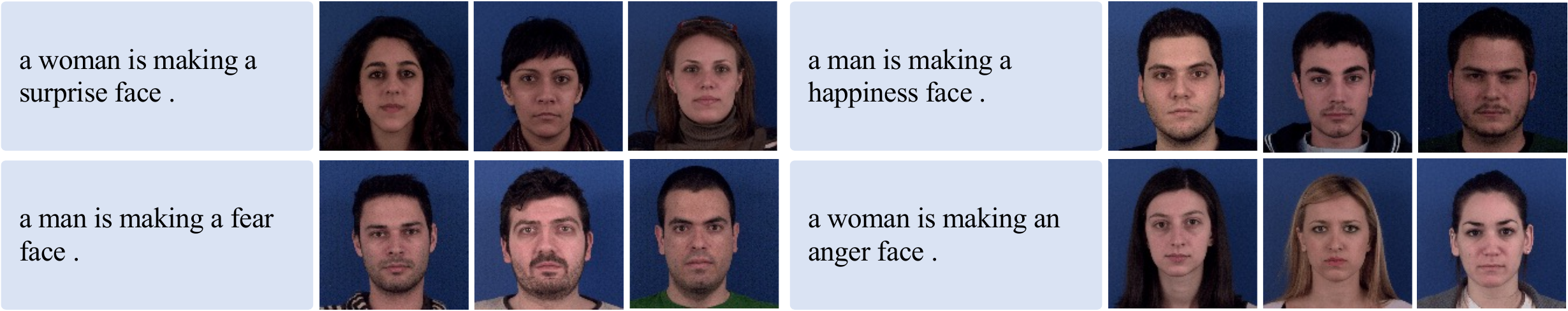}
    \caption{Example videos generated by our approach on the MUG dataset for text-to-video generation. We show three synthesized videos for each input text condition.}
    \label{fig:more_mug_our_text}
\end{minipage} \hfill
\end{figure*}

\subsection{Dataset Statistics and Textual Controls}

\noindent\textbf{Shapes}.
The Moving Shapes dataset~\cite{balaji2019conditional} includes a total of $80,640$ combinations of shape, color, size, motion, direction, and background. Almost all training videos are different as background is randomly cropped from original images.
For text-to-video and independent text-visual control experiments, we use the text descriptions provided in the original dataset. The texts are generated using a template such as \emph{``A $\langle\text{object}\rangle$ is moving in $\langle\text{motion}\rangle$ path towards $\langle\text{direction}\rangle$''} or \emph{``A $\langle\text{object}\rangle$ is moving in $\langle\text{motion}\rangle$ path in the $\langle\text{direction}\rangle$ direction''}. 
More details can be found in TFGAN~\cite{balaji2019conditional}.

\noindent\textbf{MUG}.
The MUG Facial Expression dataset~\cite{aifanti2010mug} includes $1,039$ videos with $52$ identities and $9$ motions.
The original dataset does not provide text descriptions for videos. To have a fair comparison, we follow the examples in TiVGAN~\cite{kim2020tivgan} and manually label genders for all subjects, and generate corresponding text for each video given annotations. For example, given a video with annotations as ``female'' and ``happiness'', we generate the description as \emph{``A women/young women/girl is making a happiness face''} or \emph{``A women/young women/girl is performing a happiness expression''}. We randomly choose a word to describe gender from \emph{``women'', ``young women''} and \emph{``girl''}.

\noindent\textbf{iPER}.
The iPER~\cite{liu2019liquid} dataset contains $30$ subjects wearing $103$ different clothes in total, resulting in $206$ videos (every cloth is unique in appearance and has both an A-pose and a random pose recording).
To test the generalization capability of the generation models to unseen motions, we split a held-out set of $10$ videos which contains $10$ unique appearances performing an A-pose. We further cut all videos into $100$-frame clips and perform training and evaluation on these clips. The held-out $10$ videos contain $93$ clips. Quantitative metrics are evaluated on these $93$ clips plus the same set of appearance performing a random pose ($186$ clips in total). Similar to MUG dataset, the texts are generated using a template such as \emph{``Person $\langle\text{person\_ID}\rangle$ dressed in $\langle\text{cloth\_ID}\rangle$ is performing $\langle\text{pose}\rangle$ pose''}.

\noindent\textbf{Multimodal VoxCeleb}. 
The dataset includes a total of $19,522$ videos with $3,437$ various interview situations ($453$ people).
We generate textual descriptions from annotated attributes for the Multimodal VoxCeleb dataset following previous work~\cite{xia2021tedigan}, especially this webpage\footnote{\url{https://github.com/IIGROUP/Multi-Modal-CelebA-HQ-Dataset/issues/3}}.
The attribute combinations labeled from videos of  Multimodal VoxCeleb shows a long-tail distribution (Fig.~\ref{fig:vox_stat}). There are $13,706$ unique attribute combinations out of $19,522$ samples, and $11,259$ combinations have only one data point. This motivates us to use text dropout during training as we encourage the model not to memorize certain attribute combinations with one single data point.

\begin{figure*}[h]
\begin{minipage}[t]{0.49\linewidth}
\centering
    \includegraphics[width=1\linewidth]{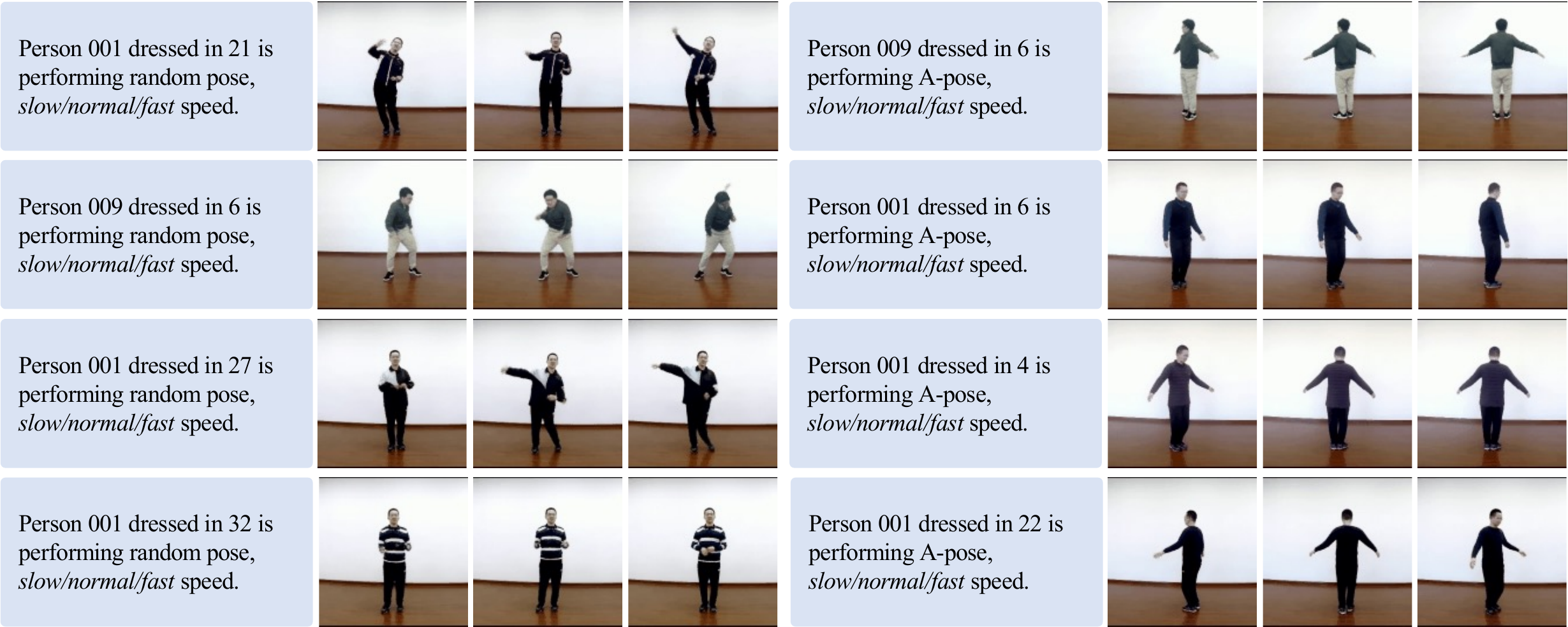}
    \caption{Example videos generated by our approach on the iPER dataset for long sequence generation. The extrapolation process is repeated for each sequence $100$ times, resulting in a $107$-frame video. The textual input also controls the speed, where ``slow'' indicates videos with slow speed such that the motion is slow, while ``fast'' indicates the performed motion is fast.  We show one synthesized video for each input text condition. The first video following the text input corresponds to the ``slow'' condition, the second corresponds to the ``normal'', and the last corresponds to the ``fast''.}
    \label{fig:more_iper_our_long}
\end{minipage} \hfill
\begin{minipage}[t]{0.49\linewidth}
    \centering
    \includegraphics[width=1\linewidth]{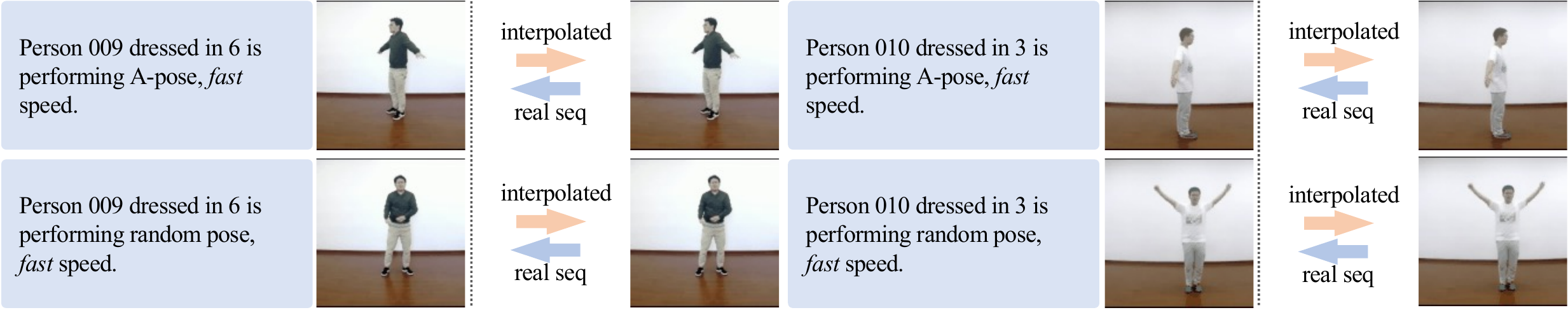}
    \caption{Example videos of our approach for video interpolation on the iPER dataset. For each example, the original real video sequence is shown on the left side and the interpolated video is shown on the right side. Interpolation is done by generating one frame between two contiguous frames.}
    \label{fig:more_iper_our_interp}
\end{minipage}\hfill
\begin{minipage}[t]{0.49\linewidth}
    \centering
    \includegraphics[width=1\linewidth]{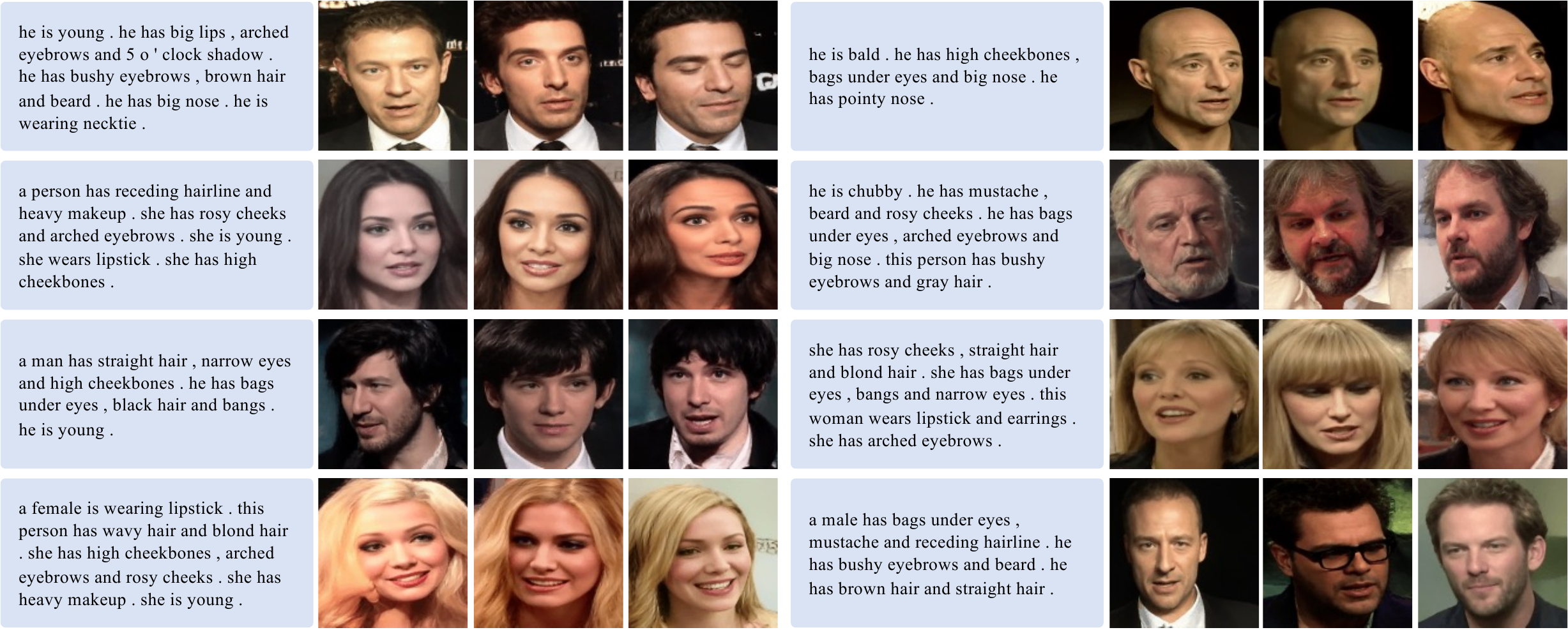}
    \caption{Example videos generated by our approach on the Multimodal VoxCeleb dataset for text-to-video generation. We show three synthesized videos for each input text condition.
    }
    \label{fig:more_vox_our_text}
\end{minipage} \hfill
\begin{minipage}[t]{0.49\linewidth}
    \centering
    \includegraphics[width=1\linewidth]{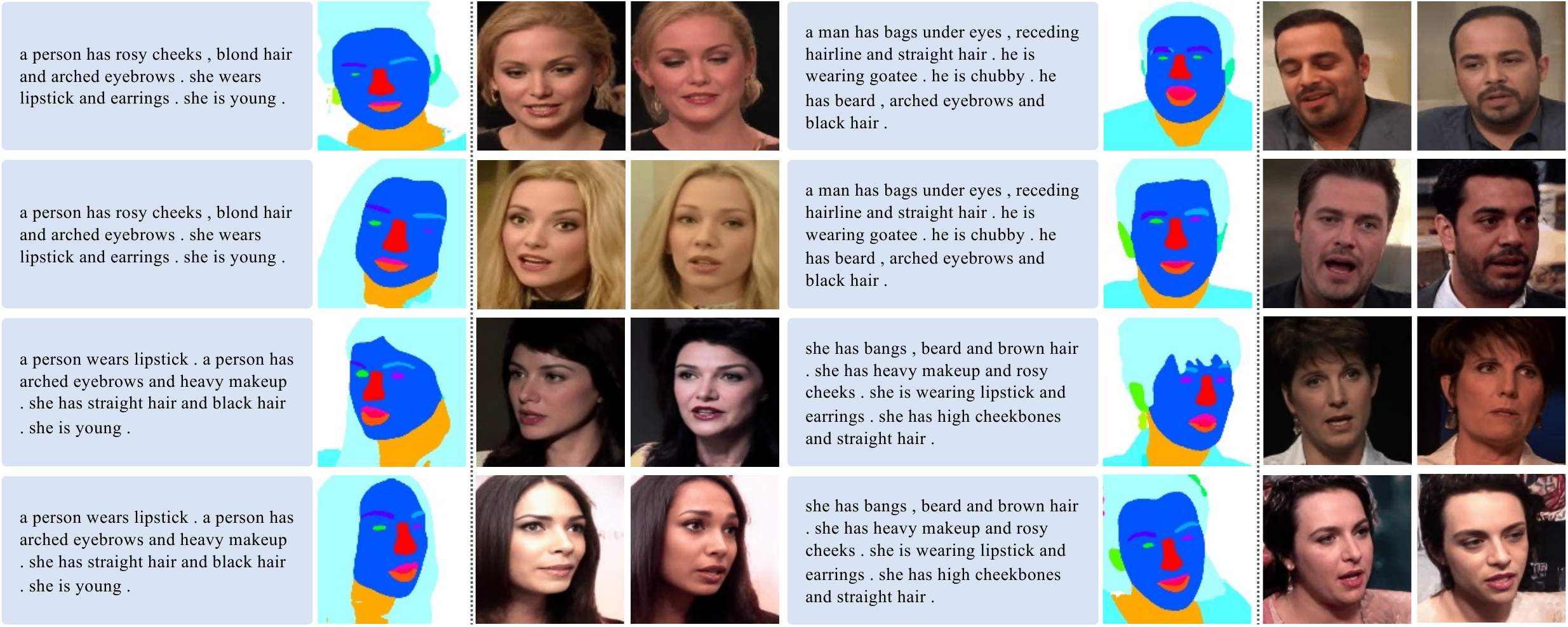}
    \caption{
    Example videos generated by our approach on the Multimodal VoxCeleb dataset for independent multimodal video generation. The input control signals are text and a segmentation mask. We show two synthesized videos for each input multimodal condition.
    }
    \label{fig:more_vox_our_text_mask}
\end{minipage} \hfill
\begin{minipage}[t]{0.49\linewidth}
    \centering
    \includegraphics[width=1\linewidth]{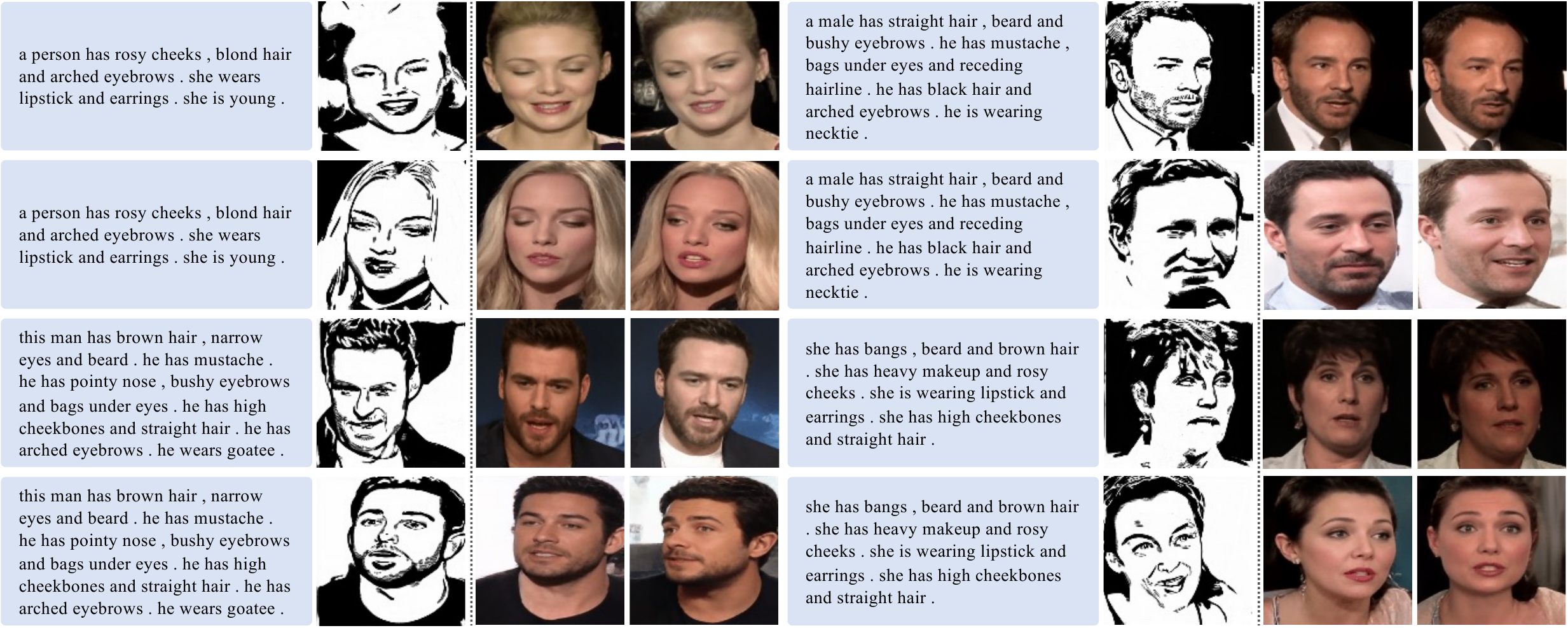}
    \caption{
    Example videos generated by our approach on the Multimodal VoxCeleb dataset for independent multimodal video generation. The input control signals are text and an artistic drawing. We show two synthesized videos for each input multimodal condition.}
    \label{fig:more_vox_our_text_draw}
\end{minipage}\hfill
\begin{minipage}[t]{0.49\linewidth}
    \centering
    \includegraphics[width=1\linewidth]{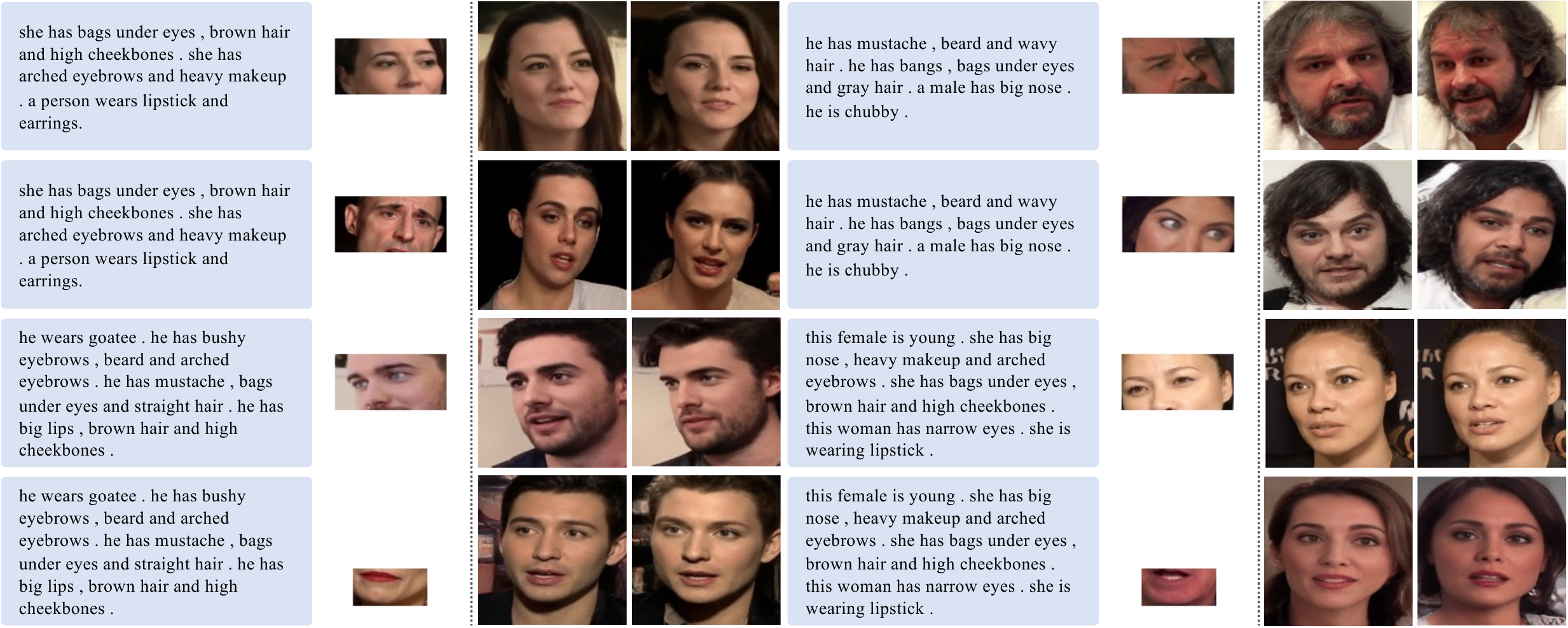}
    \caption{
    Example videos generated by our approach on the Multimodal VoxCeleb dataset for independent multimodal video generation. The input control signals are text and a partially observed image. We show two synthesized videos for each input condition.}
    \label{fig:more_vox_our_text_part}\hfill
\end{minipage}\hfill
\end{figure*}

\begin{figure*}
\begin{minipage}[t]{0.49\linewidth}
    \centering
    \includegraphics[width=1\linewidth]{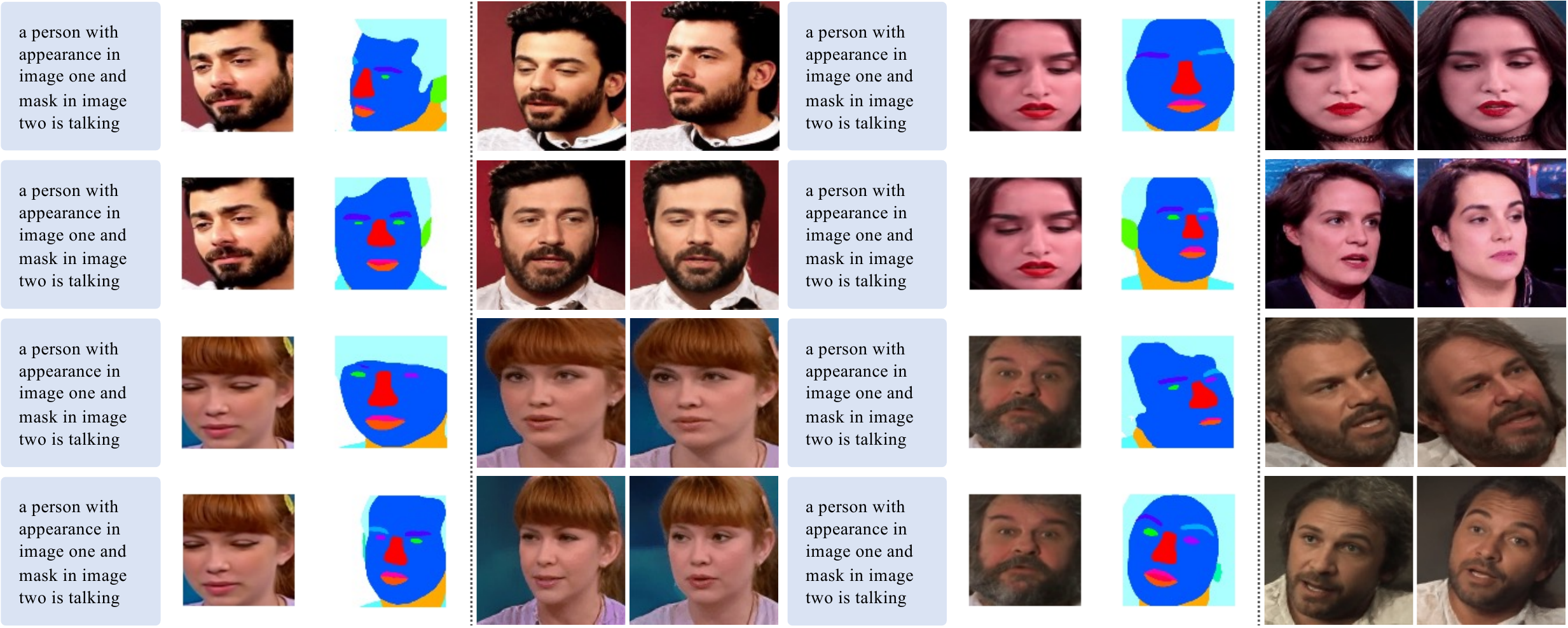}
    \caption{
    Example videos generated by our approach on the Multimodal VoxCeleb dataset for dependent multimodal video generation. The input control signals are text, an image, and a segmentation mask. We show two synthesized videos for each input condition.
    }
    \label{fig:more_vox_our_image_mask}
\end{minipage}\hfill
\begin{minipage}[t]{0.49\linewidth}
    \centering
    \includegraphics[width=1\linewidth]{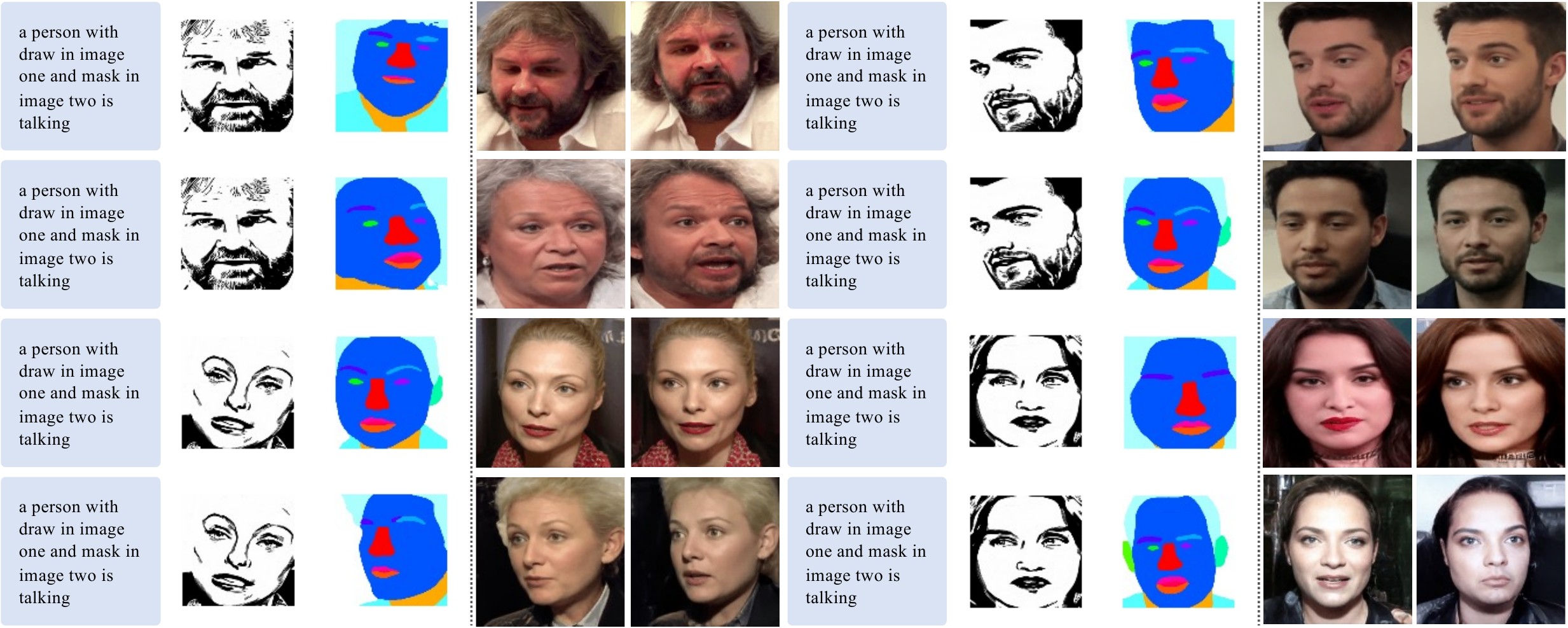}
    \caption{
    Example videos generated by our approach on the Multimodal VoxCeleb dataset for dependent multimodal video generation. The input control signals are text, an artistic drawing, and a segmentation mask. We show two synthesized videos for each input multimodal condition.}
    \label{fig:more_vox_our_draw_mask}
\end{minipage}\hfill
\begin{minipage}[t]{0.49\linewidth}
    \centering
    \includegraphics[width=1\linewidth]{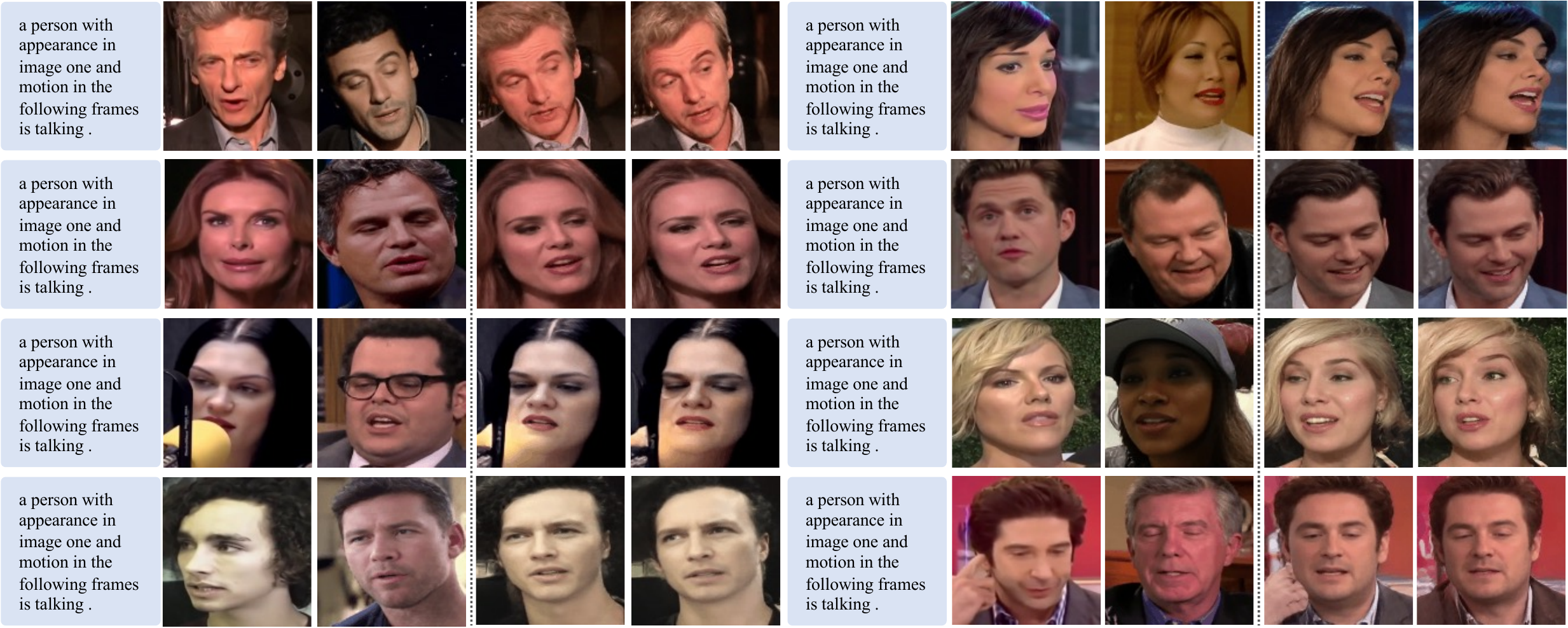}
    \caption{
    Example videos generated by our approach on the Multimodal VoxCeleb dataset for dependent multimodal video generation. The input control signals are text, an image (used for appearance), and a video (used for motion guidance, which can be better observed in our supplementary video). We show two synthesized videos for each input multimodal condition.}
    \label{fig:more_vox_our_image_video}
\end{minipage}\hfill
\begin{minipage}[t]{0.49\linewidth}
    \centering
    \includegraphics[width=1\linewidth]{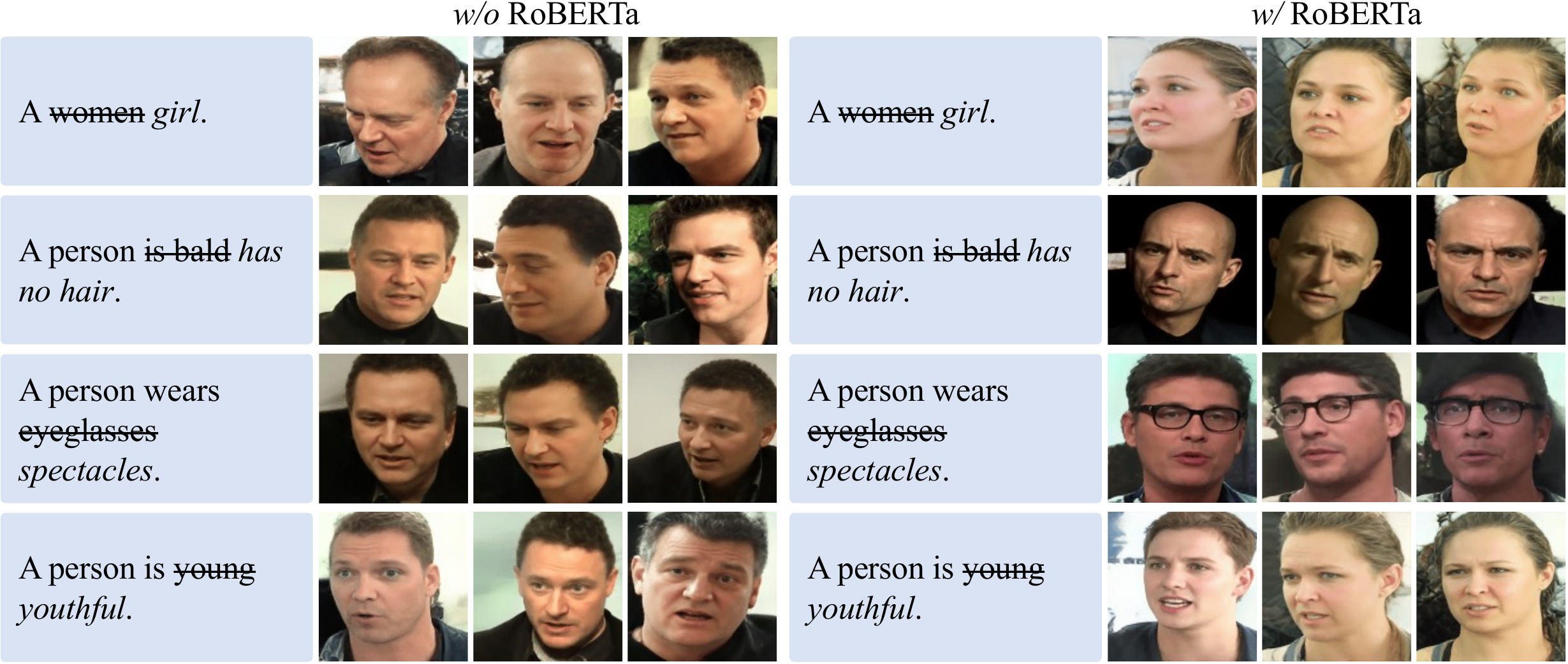}
    \caption{
    Example videos generated by methods w/ (\emph{w/} RoBERTa) and w/o (\emph{w/o} RoBERTa) using language embedding from RoBERTa as text augmentation. Models are trained on the Multimodal VoxCeleb dataset for text-to-video generation.
    We show three synthesized videos for each input text condition.
    }
    \label{fig:more_vox_our_roberta}
\end{minipage}\hfill
\begin{minipage}[t]{0.49\linewidth}
    \centering
    \includegraphics[width=1\linewidth]{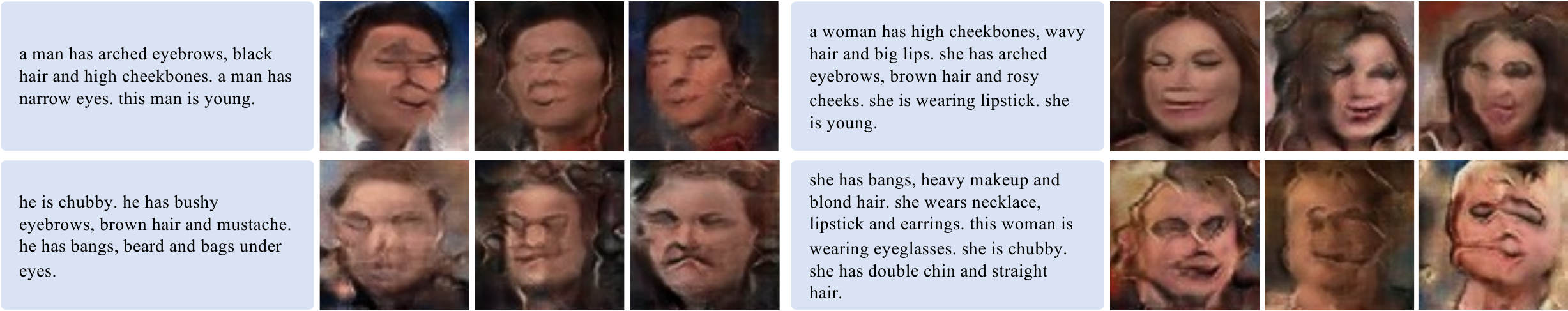}
    \caption{Example videos generated by TFGAN on the Multimodal VoxCeleb  dataset for text-to-video generation.
    We show three synthesized videos for each input text condition.}
    \label{fig:more_vox_tfgan_text}
\end{minipage}\hfill
\begin{minipage}[t]{0.49\linewidth}
    \centering
    \includegraphics[width=1\linewidth]{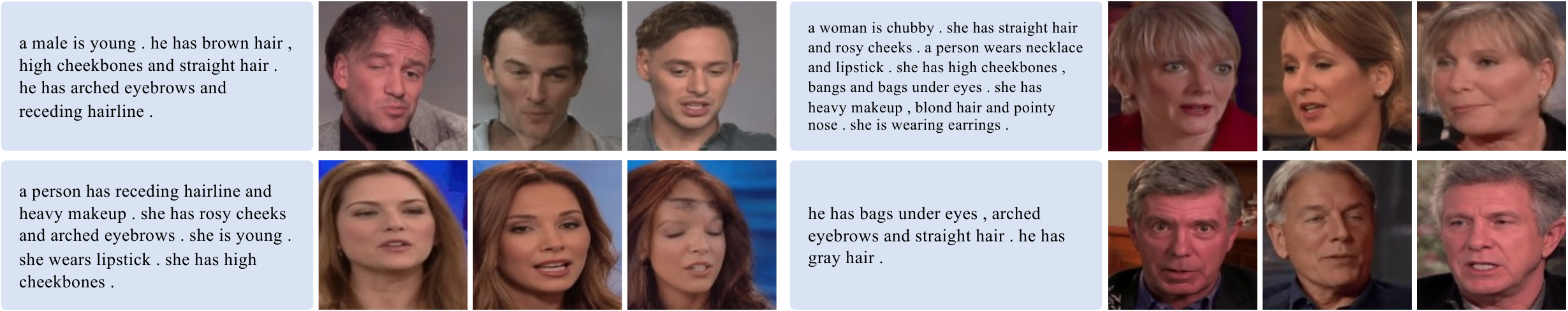}
    \caption{
    Example videos generated by ART-V on the Multimodal VoxCeleb dataset for text-to-video generation.
    We show three synthesized videos for each input text condition.}
    \label{fig:more_vox_artv_text}
\end{minipage} \hfill

\end{figure*}

\begin{figure}
\begin{minipage}[t]{1\linewidth}
    \centering
    \includegraphics[width=1\linewidth]{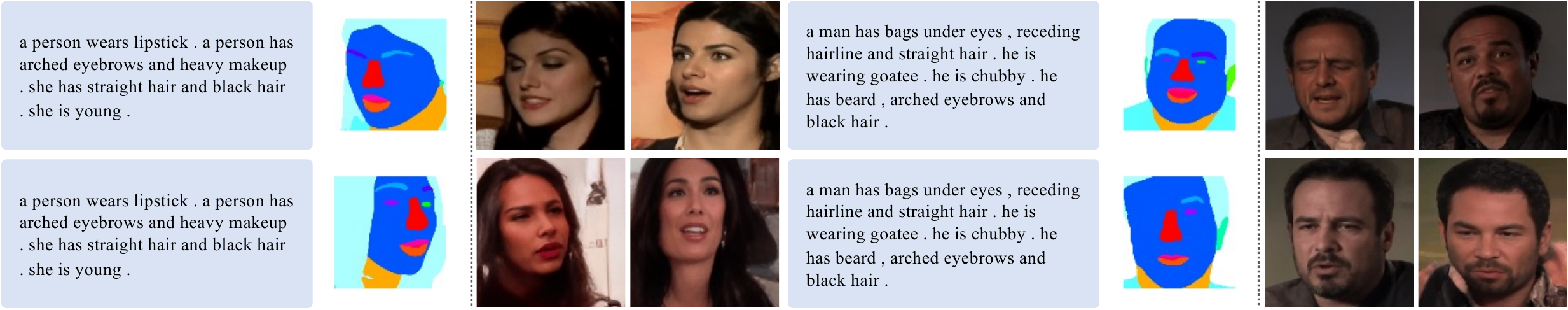}
    \caption{Example videos generated by ART-V on the Multimodal VoxCeleb dataset for independent multimodal video generation. The input control signals are text and a segmentation mask.
    We show two synthesized videos for each input multimodal condition.}
    \label{fig:more_vox_artv_text_mask}
\end{minipage}
\begin{minipage}[t]{1\linewidth}
    \centering
    \includegraphics[width=1\linewidth]{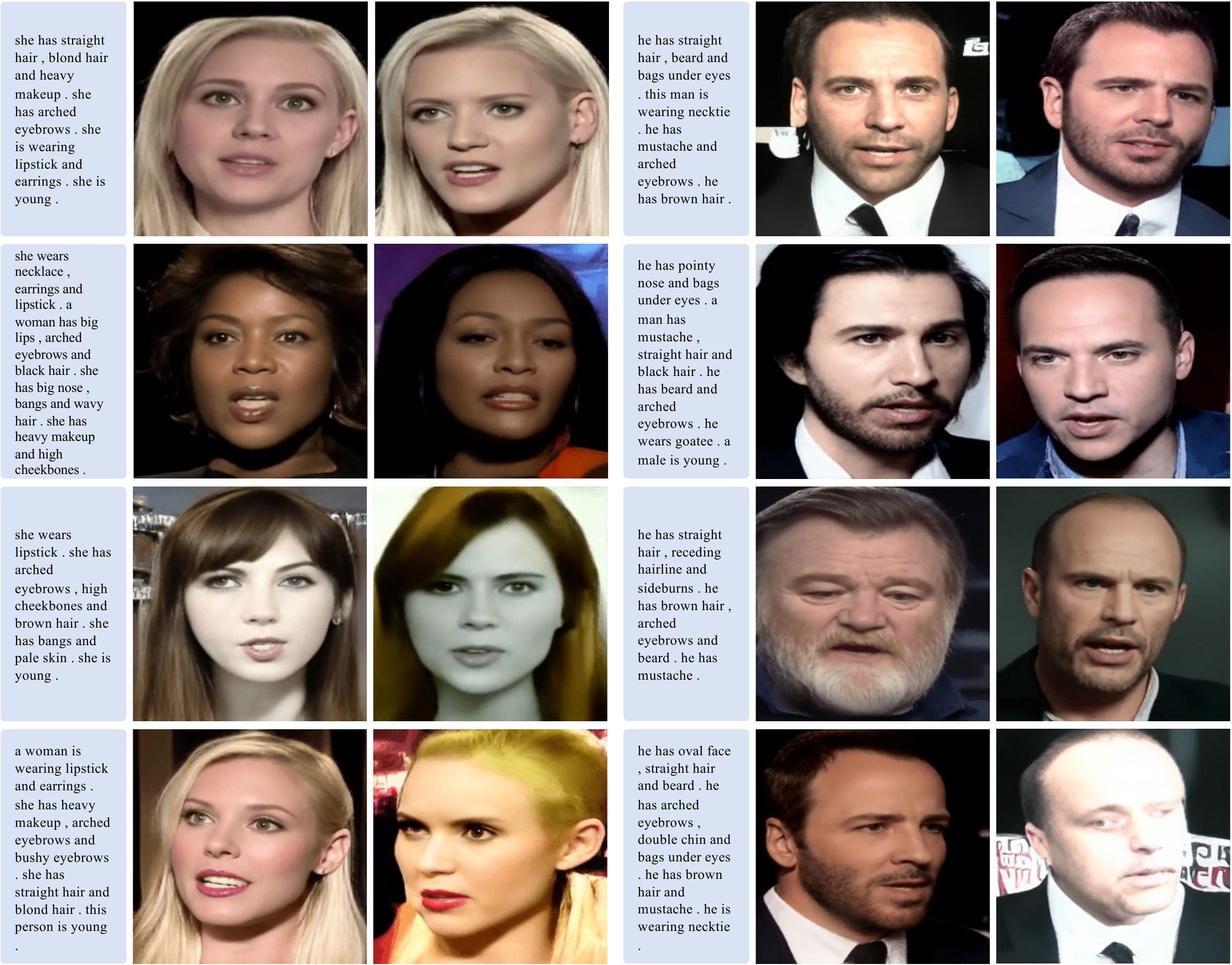}
    \caption{Example videos generated by our approach on the Multimodal VoxCeleb dataset for text-to-video generation. Videos are synthesized at a resolution of $256\times256$. We show two synthesized videos for each input text condition. We use $L=25, L_1=12$, $L_2=13$, $\alpha_1=0.9$, $\beta_1=0.1$, $\alpha_2=0.125$, and $\alpha_3=0.0625$ for mask-predict.}
    \label{fig:more_vox_our_256}
\end{minipage}\hfill
\end{figure}

\begin{figure}
\begin{minipage}[t]{1\linewidth}
    \centering
    \includegraphics[width=1\linewidth]{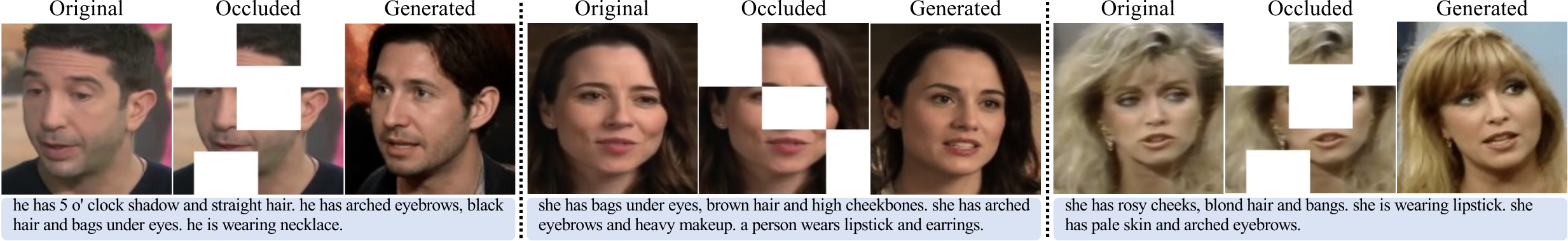}
    \caption{More example videos generated by our approach on the Multimodal VoxCeleb dataset for independent multimodal video generation. The input control signals are text and a partially observed image. Note that the occlusion pattern at test time is different from that at training time (with only either mouth or eyes and nose observable).}
    \label{fig:vox_occlude}
\end{minipage}
\begin{minipage}[t]{1\linewidth}
    \centering
    \includegraphics[width=1\linewidth]{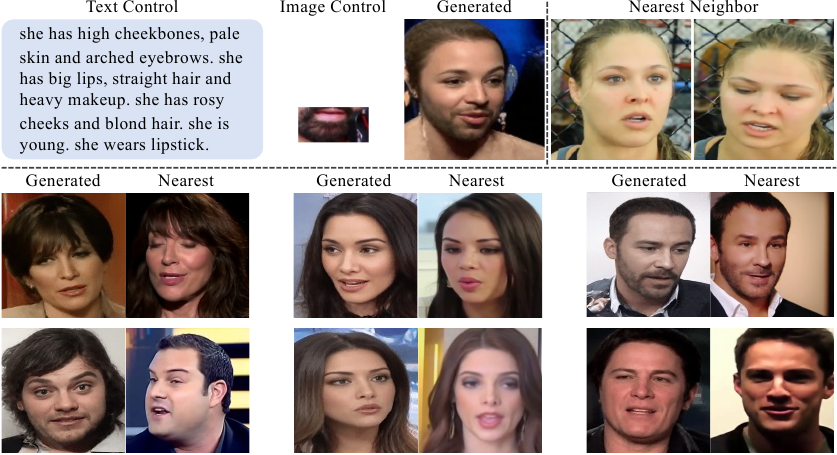}
    \caption{Examples of nearest neighbor analysis on the Multimodal VoxCeleb dataset. We show generated samples by using: 1) contradicting conditions with text suggests a female and the image shows a beard (upper, the result is an unseen combination); and 2) various conditioning (lower, conditioning omitted) with their nearest neighbors in VoxCeleb found using face similarity scores.}
    \label{fig:vox_nearest}
\end{minipage}\hfill
\end{figure}

\section{More Generated Videos}
In this section, we provide more generated videos by our approach and other works.
The thumbnail from each video is shown in the figures. Videos are in our~\href{https://snap-research.github.io/MMVID/}{\color{urlcolor}{webpage}}.

\subsection{Results on the Shapes Dataset} 

\noindent We provide more results on the Shapes dataset.
\begin{itemize}[leftmargin=1em]
  \setlength\itemsep{-0.25em}
    \item Fig.~\ref{fig:more_shape_our_text} shows the videos generated by our approach for the task of text-to-video generation.
    \item Fig.~\ref{fig:more_shape_our_text_ic} shows the videos generated by our approach for the task of independent multimodal generation.
    The input control signals are text and a partially observed image (with the center masked out).
    \item Fig.~\ref{fig:more_shape_our_depend} shows the videos generated by our approach for the task of dependent multimodal generation.
    The input control signals are text and image.
\end{itemize}

\subsection{Results on the MUG Dataset}
\noindent Fig.~\ref{fig:more_mug_our_text} shows the video generated by our approach for the task of text-to-video generation.

\subsection{Results on the iPER Dataset}
\noindent
Fig.~\ref{fig:more_iper_our_long} shows the video generated by our approach for the task of text-to-video generation. We demonstrate long sequence generation in Fig.~\ref{fig:more_iper_our_long} by performing extrapolation. The process is repeated for each sequence $100$ times, resulting in a $107$-frame video. The textual input also controls the speed, where ``slow'' indicates videos with slow speed such that the motion is slow, while ``fast'' indicates the performed motion is fast.
Fig.~\ref{fig:more_iper_our_interp} shows video interpolation results on the iPER dataset. For each example, the original real video sequence is shown on the left side, and the interpolated video is shown on the right side. Interpolation is done by generating one frame between two contiguous frames.

\subsection{Results on the Multimodal VoxCeleb Dataset} 
\noindent We provide more results for models trained on the Multimodal VoxCeleb dataset.
\begin{itemize}[leftmargin=1em]
  \setlength\itemsep{-0.25em}
    \item Fig.~\ref{fig:more_vox_our_text} shows the videos generated by our approach for the task of text-to-video generation.
    \item Fig.~\ref{fig:more_vox_our_text_mask} shows the videos generated by our approach for the task of independent multimodal generation.
    The input control signals are text and a segmentation mask.
    \item Fig.~\ref{fig:more_vox_our_text_draw} shows the videos generated by our approach for the task of independent multimodal generation.
    The input control signals are text and an artistic drawing.
    \item Fig.~\ref{fig:more_vox_our_text_part} shows the videos generated by our approach for the task of independent multimodal generation.
    The input control signals are text and a partially observed image.
    \item Fig.~\ref{fig:more_vox_our_image_mask} shows the videos generated by our approach for the task of dependent multimodal generation.
    The input controls are text, an image, and a segmentation mask.
    \item Fig.~\ref{fig:more_vox_our_draw_mask} shows the videos generated by our approach for the task of dependent multimodal generation.
    The input control signals are text, an artistic drawing, and a segmentation mask.
    \item Fig.~\ref{fig:more_vox_our_image_video} shows the videos generated by our approach for the task of dependent multimodal generation.
    The input control signals are text, an image (used for appearance),  and a video (used for motion guidance, which can be better observed in our supplementary video).
    \item Fig.~\ref{fig:more_vox_our_roberta} shows the videos generated by methods w/ (\emph{w/} RoBERTa) and w/o (\emph{w/o} RoBERTa) using language embedding from RoBERTa~\cite{liu2019roberta} as text augmentation. Models are trained on the Multimodal VoxCeleb dataset for text-to-video generation.
    \item Fig.~\ref{fig:more_vox_tfgan_text}, Fig.~\ref{fig:more_vox_artv_text}, and Fig.~\ref{fig:more_vox_artv_text_mask} show the videos synthesized by TFGAN for text-to-video generation, ART-V for text-to-video generation, and ART-V for independent multimodal generation, respectively. Artifacts can be observed from the generated videos.
    \item Fig.~\ref{fig:vox_occlude} shows more videos generated from partially occluded faces. Note that the occlusion pattern at test time is different from that at training time.
    \item Fig.~\ref{fig:vox_nearest} shows examples of nearest neighbor analysis on the Multimodal VoxCeleb dataset. We show generated samples by using: 1) contradicting conditions with text suggests a female and the image shows a beard (upper, the result is an unseen combination); and 2) various conditioning (lower, conditioning omitted) with their nearest neighbors in VoxCeleb found using face similarity scores\footnote{Face Recognition Code:~\url{ https://github.com/timesler/facenet-pytorch}}.
\end{itemize}

\section{Limitation and Future Work}
\noindent\textbf{Higher Resolution Generation}.
We conduct experiments to generate higher resolution videos by performing experiments on the Multimodal VoxCeleb dataset to synthesize video with the resolution of $256\times256$. Synthesized videos for the task of text-to-video generation are shown in Fig.~\ref{fig:more_vox_our_256}. We notice that artifacts can be found from some videos: \eg we find synthesized samples are more likely to show weird colors (Fig.~\ref{fig:more_vox_our_256}, the third row and the second sample) or appear to be blurry (Fig.~\ref{fig:more_vox_our_256} the last row and the second and the fourth sample). We also find that the temporal consistency is worse than videos generated at the resolution of $128\times128$. One possible reason might be that each frame at a higher resolution requires a longer token sequence. Therefore, an image-level temporal regularization constraint might be necessary to improve the video consistency, which we leave for future work.

\noindent\textbf{Longer Sequence Generation}. For the task of long sequence generation, we notice that extrapolation might not always give reasonable motion patterns. For example, as shown in Fig.~\ref{fig:more_iper_our_long}, the textual inputs from the first row struggle to generate diverse motions when the speed is given as ``slow''. This might be due to the temporal step size used to sample frames during training being short when the speed is ``slow'' and the frames cannot always cover diverse motion patterns.  A future direction could be balancing the training set to cover enough motion patterns for the sampled frames. 

\noindent\textbf{Diversity of Non-Autoregressive Transformer}. Compared with the autoregressive transformer, the non-autoregressive transformer can generate videos with better temporal consistency. However, we notice that the autoregressive transformer might generate more diverse videos, though many have low video quality. We apply text augmentation and improved mask-predict to improve the diversity of the non-autoregressive transformer. An interesting research direction is how to unify the training methods from the autoregressive and non-autoregressive transformer to enhance the non-autoregressive transformer itself for generating more diverse videos. We leave the direction as to future work.

\section{Ethical Implications}
Our method can synthesize high-quality videos with multimodal inputs. However, a common concern among the works for high fidelity and realistic image and video generation is the purposely abusing the technology for nefarious objectives. The techniques developed for synthetic image and video detection can help alleviate such a problem by automatically finding artifacts that humans might not easily notice.

\end{document}

%% file: math_commands.tex
\usepackage{amsmath,amsfonts,bm}

\def\eqref#1{equation~\ref{#1}}

\def\1{\bm{1}}

\DeclareMathAlphabet{\mathsfit}{\encodingdefault}{\sfdefault}{m}{sl}
\SetMathAlphabet{\mathsfit}{bold}{\encodingdefault}{\sfdefault}{bx}{n}

\DeclareMathOperator*{\argmax}{arg\,max}